\documentclass{article}

\usepackage[main, final]{neurips_2026}
\usepackage{array}

\usepackage[utf8]{inputenc} %
\usepackage[T1]{fontenc}    %
\usepackage{hyperref}       %
\usepackage{url}            %
\usepackage{booktabs}       %
\usepackage{amsfonts}       %
\usepackage{nicefrac}       %
\usepackage{microtype}      %
\usepackage{xcolor}         %

\usepackage{amsmath}
\usepackage{amssymb}
\usepackage{amsthm}
\usepackage{multirow}
\usepackage{algorithm}
\usepackage{algpseudocode}
\usepackage{bm}
\usepackage{cuted}
\usepackage{capt-of}
\usepackage{listings}
\usepackage{wrapfig}
\usepackage{subcaption}
\usepackage{tabularx}
\usepackage[export]{adjustbox}
\usepackage{soul}
\usepackage{placeins}
\hypersetup{hidelinks}
\theoremstyle{plain}
\newtheorem{theorem}{Theorem}
\newtheorem{proposition}{Proposition}
\newtheorem{lemma}{Lemma}
\newtheorem{corollary}{Corollary}

\theoremstyle{definition}
\newtheorem{assumption}{Assumption}

\theoremstyle{remark}

\providecommand{\pvz}{\mathbf{z}}
\providecommand{\pvu}{\mathbf{u}}
\providecommand{\pvv}{\mathbf{v}}
\providecommand{\pvx}{\mathbf{x}}
\providecommand{\pvy}{\mathbf{y}}
\providecommand{\pvg}{\mathbf{g}}
\providecommand{\pvn}{\mathbf{n}}
\providecommand{\pvmu}{\boldsymbol{\mu}}
\providecommand{\pveps}{\boldsymbol{\epsilon}}
\providecommand{\pvDelta}{\boldsymbol{\Delta}}
\providecommand{\pvN}{\mathcal{N}}
\providecommand{\pvKL}{\mathrm{KL}}

\title{Anchored Inertial Proposals with Objective Verification for Diffusion-Prior Inverse Problems}

\author{%
  Minwoo Kim\thanks{These authors contributed equally to this work} \quad
  Seunghyeok Shin\footnotemark[1] \quad
  Dabin Kim \quad
  Hongki Lim\textsuperscript{$\dagger$}\\
  Department of Electrical and Computer Engineering, Inha University\\
  Incheon 22212, South Korea\\
  \texttt{\{ququlza1520,ssh8642,1124db\}@inha.edu}, \texttt{hklim@inha.ac.kr}
}

\begin{document}

\maketitle

\begingroup
\renewcommand{\thefootnote}{\fnsymbol{footnote}}
\footnotetext[2]{Corresponding author.}
\endgroup

\begin{abstract}
Training-free diffusion priors are powerful for inverse problems, but measurement guidance during reverse sampling is local: aggressive updates can improve immediate data fit while disrupting later denoising. We introduce an anchored inertial correction rule that separates proposal generation from acceptance. At each corrected step, the denoiser prediction anchors a frozen clean-space objective combining measurement fit with a noise-level-dependent proximal term. Heavy-ball dynamics with step size and momentum computed from the measured curvature generate a clean-state candidate; a separate verifier evaluates it with the same objective, with the denoiser prediction retained as fallback; both stages are derived from one objective for the whole trajectory, considered one step at a time. The accepted clean state is re-noised to continue sampling. The rule applies to pixel and latent diffusion with differentiable measurement operators, requires no retraining, and adds no denoiser calls inside the correction loop. We analyze how objective verification yields local nonincrease and displacement control. Across natural-image benchmarks, the method achieves competitive reconstruction quality. Code is available at \href{https://github.com/Seunghyeok0715/PIVOT2026}{\nolinkurl{github.com/Seunghyeok0715/PIVOT2026}}.
\end{abstract}
\section{Introduction}

Image reconstruction from degraded measurements is a canonical inverse problem: blur, masking, downsampling, Fourier-magnitude sensing, nonlinear camera response, and related forward models can make recovery from noisy observations nonunique or unstable without assumptions on the image class \cite{hadamard1923cauchy,tikhonov1963solution,engl1996regularization}. Variational and Bayesian methods impose such assumptions through data fidelity and prior regularization \cite{tarantola2005inverse,kaipio2005statistical}; total variation and sparsity are classic examples in imaging and compressed sensing \cite{rudin1992nonlinear,donoho2006compressed,candes2006robust}. Deep generative priors instead encode image structure through learned or architectural models, including generator-range constraints, implicit network priors, and robust generative-prior formulations \cite{bora2017compressed,ulyanov2018deep,jalal2020robust}.

\begin{figure}
\vspace{1em}
    \centering
    \includegraphics[width=\linewidth]{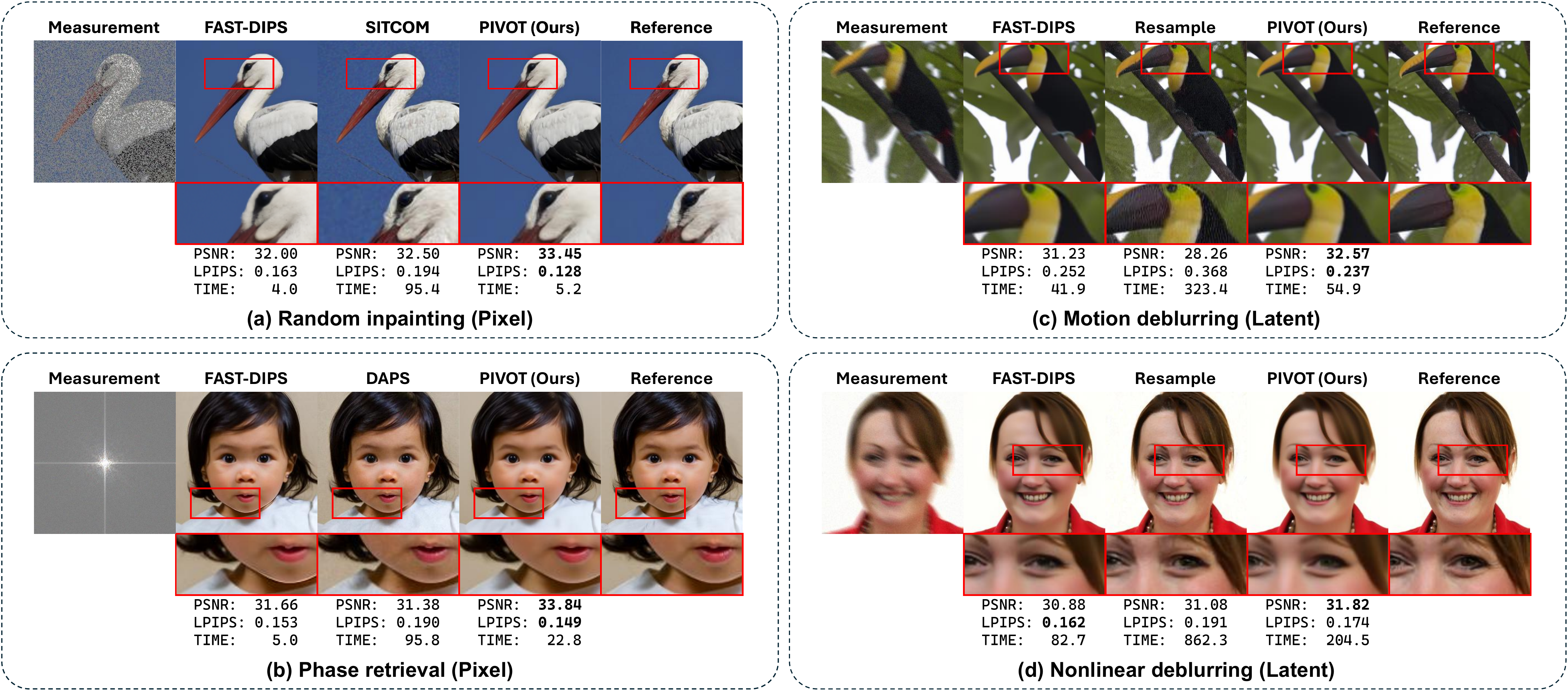}
    \vspace{1em}
    \caption{
\textbf{Representative reconstructions.}
Qualitative comparison on pixel-space random inpainting and Phase retrieval, and latent-space motion and nonlinear deblurring. Red boxes show zoomed regions. Metrics are per-image PSNR$\uparrow$, LPIPS$\downarrow$, and sampling time$\downarrow$. \textsc{PIVOT} preserves sharper details with favorable speed--quality trade--offs.
}
\label{fig:qualitative_reconstructions}
\end{figure}

Diffusion and score-based models are attractive for inverse problems because they provide denoising information across noise levels rather than a single static generator. This line builds on score matching, denoising score matching, diffusion-based generative modeling, and continuous-time score formulations \cite{hyvarinen2005estimation,vincent2011connection,sohl2015deep,song2019generative,ho2020denoising,song2021score}. Modern samplers and parameterizations have made diffusion models strong image priors \cite{nichol2021improved,dhariwal2021diffusion,karras2022elucidating}, while latent diffusion enables high-resolution generation by denoising in a compressed latent space and decoding to images \cite{rombach2022high}. For reconstruction, this suggests training-free conditioning: use a pretrained diffusion model as an image prior while injecting measurement information during the reverse process.

The difficulty is that measurement correction in a reverse diffusion trajectory is local but consequential. At each step, the denoiser produces a clean prediction from the current noisy state, and a conditioning method must decide how to modify this prediction using the measurement. A correction that improves a local guidance signal can still perturb later denoising, especially under nonlinear measurements, robust losses, high-noise states, or latent decoder geometry. Existing training-free solvers construct such corrections using likelihood guidance, manifold or projection constraints, pseudoinverse updates, posterior approximations, plug-and-play corrections, or operator-specific structure \cite{song2022solving,chung2022improving,kawar2022denoising,wang2023zero,meng2022diffusion,chung2023diffusion,song2023pseudoinverse,zhu2023denoising,rout2023solving}. Recent work further develops decoupled conditioning principles by separating data consistency from denoising or noise evolution, formulating variational or annealed posterior corrections, enforcing hard data consistency in latent diffusion, using consistency-model guidance, or reusing gradient history and momentum \cite{song2024solving,chung2024decomposed,mardani2024variational,zhang2025improving,song2023consistency,zhao2024cosign,li2024decoupled,he2023fast,wizadwongsa2024diffusion}. Our focus is a finer decoupling within each corrected reverse step: separating proposal generation from the decision to commit the proposal to the trajectory.

We introduce \emph{Propose with Inertia, Verify with an Objective Test} (\textsc{PIVOT}), a training-free correction rule for diffusion-prior inverse
problems with differentiable measurement operators. At each corrected reverse
step, \textsc{PIVOT} anchors a frozen clean-space objective at the denoiser
prediction, combining measurement fit with a noise-level-dependent quadratic
proximal term. This objective provides a local criterion for deciding whether a clean
correction should be committed to the reverse trajectory. The interpretation is
deliberately local: the objective checks corrections at the current step, but
does not define a global MAP objective, exact posterior, or calibrated posterior
sampler.

\textsc{PIVOT} separates proposal generation from acceptance. The proposal
branch uses a small number of heavy-ball iterations to generate
a candidate clean state {\cite{polyak1964some}}, with step size and momentum computed at every step from the curvature measured along the iterates. The verification branch evaluates the final iterate and the denoiser prediction, and selects the candidate with the lower
value under the frozen objective. In short, inertia proposes; the frozen objective verifies. Both branches are derived from one objective for the whole trajectory whose per-step terms are the frozen objectives, each with a known weight, considered one step at a time: the proposal is the heavy-ball method matched to the curvature of the frozen objective as estimated from gradient secants, and verification commits a correction only if it lowers the frozen objective below its value at the denoiser prediction. The accepted clean state is
then reinserted into the reverse trajectory by fresh noising at the next
scheduled noise level. The same rule applies to pixel and latent diffusion: in
the latter, the clean variable is decoded before applying the measurement
operator, so gradients pass through both the decoder and the forward model.

Our contributions are:
\begin{itemize}
    \item A training-free correction rule based on anchored inertial
    proposals with objective verification. At each corrected reverse step,
    heavy-ball dynamics with step size and momentum computed from the measured curvature generate a clean-space candidate,
    while a denoiser-anchored objective accepts or rejects the candidate, with the denoiser prediction retained as fallback.

    \item A local analysis of the proposal--verification interface. We derive the frozen objective, the proposal and the verifier from one trajectory objective considered one step at a time, and characterize
    how
    objective verification yields frozen-objective nonincrease
    and displacement control under stated local assumptions.

        \item A unified pixel- and latent-diffusion instantiation for differentiable
    measurement operators, requiring no retraining and no additional denoiser
    evaluations inside the correction loop. Experiments on FFHQ and ImageNet demonstrate competitive reconstruction quality{}, and component ablations show gains from the heavy-ball proposal rule over gradient steps and from the conservative first step, while the effect of the curvature policies depends on the task.
\end{itemize}
\section{Background}

\subsection{Problem setting and diffusion anchor}

We consider inverse problems with observations \(\mathbf{y}=A(\mathbf{x}_\star)+\boldsymbol{\eta}\), where \(\mathbf{x}_\star\) is the unknown image, \(A\) is a known measurement operator, and \(\boldsymbol{\eta}\) models noise or mismatch. The operator \(A\) may be linear or nonlinear, but is assumed differentiable wherever gradients are used; a clip at its output instead receives the gated derivative of Section~\ref{sec:method_objective_smoothing}. Because such problems are generally ill posed without image prior information \cite{hadamard1923cauchy,tikhonov1963solution,engl1996regularization}, reconstruction is often written as \(\min_{\mathbf{x}}\ell(A(\mathbf{x}),\mathbf{y})+\lambda R(\mathbf{x})\), with measurement loss \(\ell\) and prior or regularizer \(R\) \cite{tarantola2005inverse,kaipio2005statistical}. Here, the prior is accessed through a pretrained diffusion clean predictor \(D_\sigma(\mathbf{u})\), which maps a noisy state \(\mathbf{u}\) at noise level \(\sigma\) to a clean estimate \cite{ho2020denoising,song2021score,karras2022elucidating}. Along the reverse trajectory, \(\mathbf{u}_k\) and \(\sigma_k\) denote the current state and noise level, and \(\boldsymbol{\mu}_k=D_{\sigma_k}(\mathbf{u}_k)\) serves as the stepwise clean anchor. We do not assume that this prediction provides calibrated posterior uncertainty or an exact covariance.

\subsection{Pixel and latent reconstruction variables}

We write both pixel and latent diffusion in terms of a clean model-space variable \(\mathbf{z}\). Its image-space map is \(G(\mathbf{z})=\mathbf{z}\) for pixel diffusion and \(G(\mathbf{z})=\mathrm{Dec}(\mathbf{z})\) for latent diffusion, where \(\mathrm{Dec}\) is the pretrained latent-diffusion decoder \cite{rombach2022high}. The predicted measurement is \(F(\mathbf{z})=A(G(\mathbf{z}))\). Thus the same notation covers image-space reconstruction and decoder-composed measurements. For differentiable \(A\) and \(G\), measurement gradients are pulled back to \(\mathbf{z}\) through
\[
    \nabla_{\mathbf{z}} \ell(F(\mathbf{z}),\mathbf{y})
    =
    J_G(\mathbf{z})^\top
    J_A(G(\mathbf{z}))^\top
    \nabla_{\hat{\mathbf{y}}}\ell(\hat{\mathbf{y}},\mathbf{y}),
    \qquad
    \hat{\mathbf{y}}=A(G(\mathbf{z})).
\]
This differentiable composite map is the main requirement for the local correction rule below.

\subsection{Local correction primitives}

At a fixed reverse step, the denoiser prediction defines a local objective of the form \(\ell(F(\mathbf{z}),\mathbf{y})+\frac{\beta_k}{2}\|\mathbf{z}-\boldsymbol{\mu}_k\|_2^2\). The quadratic term is a proximal penalty that keepsthe clean variable near the current denoiser prediction; it is a stepwise surrogate, not a global image prior. The method uses two standard local primitives.  First, heavy-ball dynamics provide finite-step inertial updates for candidate generation \cite{polyak1964some}, with step size and momentum set from curvature estimated by gradient secants \cite{barzilai1988two}. Second, verification evaluates the candidate on the frozen objective and retains it only if it lowers this objective, with the denoiser prediction available as fallback \cite{armijo1966minimization,nocedal2006numerical}. These ingredients provide the technical scaffold for \textsc{PIVOT}: a differentiable measurement loss, a denoiser-centered anchor, the composite map \(F=A\circ G\), and local primitives for proposal generation and objective-based verification.

\section{Method}
\label{sec:method}

\begin{figure}[t]
    \centering
    \includegraphics[width=\linewidth]{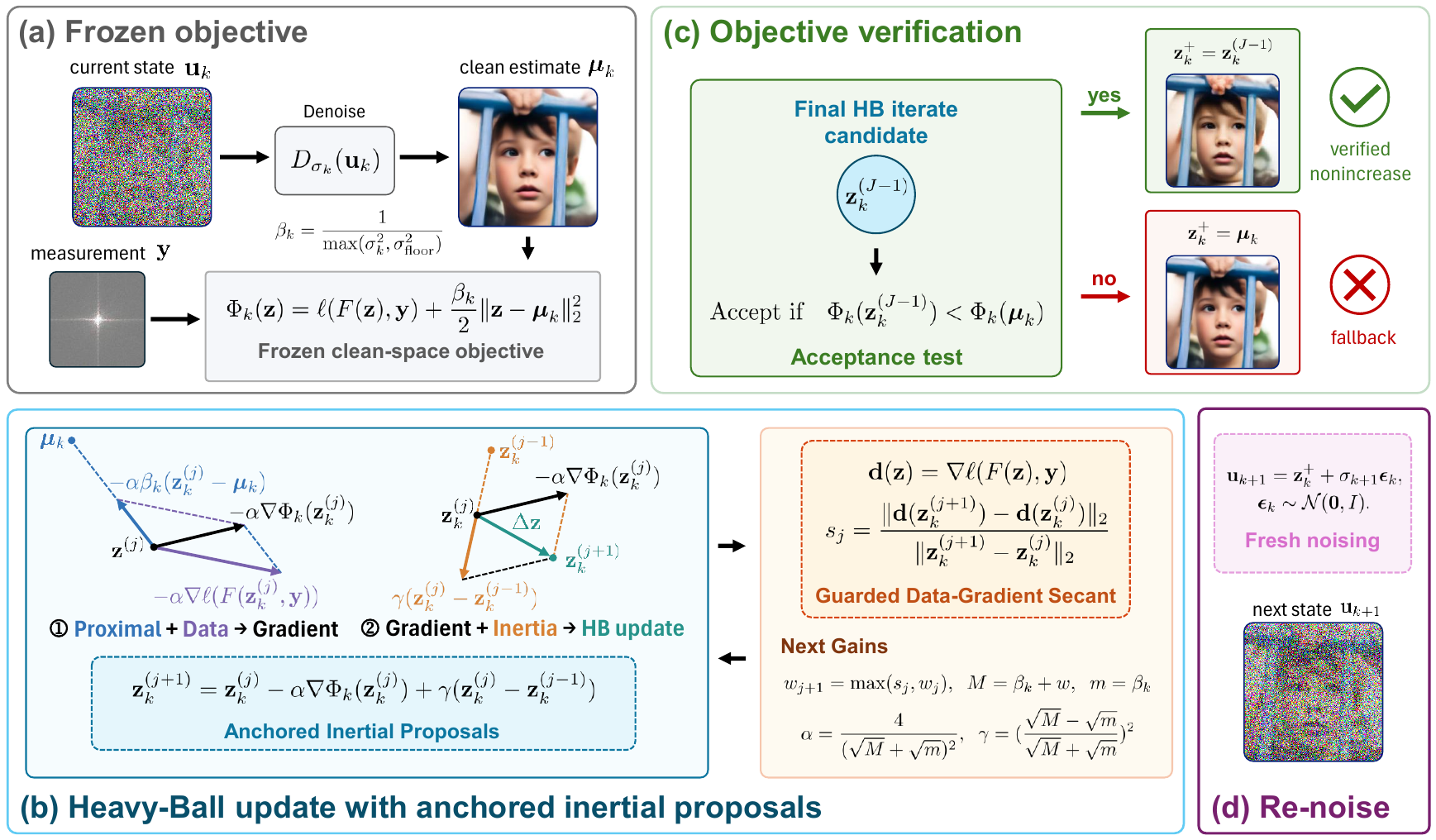}
\caption{
\textbf{\textsc{PIVOT} overview. }(a) The denoiser prediction \(\boldsymbol{\mu}_k\) anchors a frozen clean-space objective \(\Phi_k\) combining measurement fidelity with a noise-dependent proximal term. (b) Starting from \(\boldsymbol{\mu}_k\), heavy-ball (HB) updates combine data and proximal term gradients with inertia; step size and momentum are computed from the proximal term curvature and guarded data-gradient secant estimates. (c) The final HB iterate is accepted only if it lowers the same frozen objective relative to \(\boldsymbol{\mu}_k\); otherwise, the anchor is retained, ensuring \(\Phi_k(\mathbf{z}_k^+)\leq\Phi_k(\boldsymbol{\mu}_k)\). (d) Fresh Gaussian noise at the next scheduled noise level is added to the selected clean state to continue reverse diffusion. The same procedure applies in pixel or latent space without additional denoiser evaluations inside the correction loop.
}
    \label{fig:proposal_verification}
\end{figure}

\nopagebreak\paragraph{Motivation and main idea.}
A measurement correction inside reverse sampling should improve the data fit without moving the state away from what the prior predicts; a single guidance step trades one against the other through its scale. \textsc{PIVOT} separates the two roles: a fast local optimizer proposes a correction, and a fixed criterion decides whether to keep it. The method follows from this idea in three steps. First, charging each correction for the information it adds to the reverse chain makes the criterion the frozen objective \(\Phi_k\): the data misfit plus a proximal term around the denoiser prediction (Section~\ref{sec:method_overview}). Second, the proposal is a few heavy-ball steps on \(\Phi_k\), with step size and momentum computed from curvature measured along the iterates rather than tuned (Section~\ref{sec:method_proposal_verification}). Third, the proposal is kept only if it lowers \(\Phi_k\) below its value at the denoiser prediction, which bounds every accepted displacement (Section~\ref{sec:method_theory}); the kept state is then re-noised.

\subsection{Overview and design principle}
\label{sec:method_overview}

At reverse step \(k\), the denoiser produces \(\boldsymbol{\mu}_k=D_{\sigma_k}(\mathbf{u}_k)\). \textsc{PIVOT} uses \(\boldsymbol{\mu}_k\) as a clean anchor and corrects the clean variable \(\mathbf{z}\), with measurement prediction \(F(\mathbf{z})=A(G(\mathbf{z}))\). From a posterior-sampling perspective, the Gaussian likelihood is \(p(\mathbf{y}\mid\mathbf{z})\propto\exp\big(-\|F(\mathbf{z})-\mathbf{y}\|_2^2/(2\sigma_y^2)\big)\). Since the diffusion prior is available only implicitly through denoising, \textsc{PIVOT} uses the local surrogate \(\tilde p_k(\mathbf{z}\mid\mathbf{u}_k)=\mathcal{N}(\mathbf{z};\boldsymbol{\mu}_k,\beta_k^{-1}I)\). Combining the likelihood and surrogate gives
\[
\begin{aligned}
    \pi_k(\mathbf{z}\mid\mathbf{u}_k,\mathbf{y})
    &\propto p(\mathbf{y}\mid\mathbf{z})\tilde p_k(\mathbf{z}\mid\mathbf{u}_k) \\
    &\propto
    \exp\{-\frac{1}{2\sigma_y^2}\|F(\mathbf{z})-\mathbf{y}\|_2^2\}
    \exp\!\left\{-\frac{\beta_k}{2}\|\mathbf{z}-\boldsymbol{\mu}_k\|_2^2\right\}.
\end{aligned}
\]
The negative log, up to constants independent of \(\mathbf{z}\), is the frozen objective
\begin{equation}
    \Phi_k(\mathbf{z})
    =
    \ell(F(\mathbf{z}),\mathbf{y})
    +
    \frac{\beta_k}{2}
    \|\mathbf{z}-\boldsymbol{\mu}_k\|_2^2 .
    \label{eq:frozen_objective}
\end{equation}
where \(\ell(F(\mathbf{z}),\mathbf{y})=\|F(\mathbf{z})-\mathbf{y}\|_2^2/(2\sigma_y^2)\) is the Gaussian negative log-likelihood and \(\sigma_y\) is the likelihood scale (Appendix~\ref{app:exp_detail}). This interpretation is local: \(\Phi_k\) checks corrections around the current denoiser prediction, but does not define a global MAP objective, exact posterior, or exact posterior sampler.

The same objective is the per-step term of one objective for the whole reverse trajectory. Let \(Q\) be the law of the states \(\mathbf{v}_{0:K-1}\) of the uncorrected chain \(\mathbf{v}_{k+1}=D_{\sigma_k}(\mathbf{v}_k)+\sigma_{k+1}\boldsymbol{\epsilon}_k\) and \(P\) that of the states \(\mathbf{u}_{0:K-1}\) of the corrected chain \(\mathbf{u}_{k+1}=\mathbf{z}_k^++\sigma_{k+1}\boldsymbol{\epsilon}_k\), from the same initial law. The weighted fit to the measurements along the trajectory plus the information that the corrections add relative to the uncorrected reverse chain decomposes over the steps (Corollary~\ref{cor:trajectory_objective}, with \(\beta_k=\sigma_k^{-2}\)):
\begin{equation}
    \begin{aligned}
    \mathcal{J}(P)&=\mathbb{E}_P\Big[\sum_{k\le K-2}\lambda_k\,\ell(F(\mathbf{z}_k^+),\mathbf{y})\Big]+\mathrm{KL}(P\,\|\,Q)\\
    &=\sum_{k\le K-2}\lambda_k\,\mathbb{E}_P\big[\Phi_k(\mathbf{z}_k^+)\big],\qquad \lambda_k=\sigma_k^2/\sigma_{k+1}^2 .
    \end{aligned}
    \label{eq:trajectory_objective}
\end{equation}
Choosing each correction to minimize its own term, given the past and without modelling its effect on later anchors, gives the per-step problem \(\min_{\mathbf{z}}\Phi_k(\mathbf{z})\) (the last step, which has no reinsertion noise, uses the same problem); the proximal term is the information price of moving away from the anchor. \null

A small number of heavy-ball iterations on \(\Phi_k\) generates a candidate correction \cite{polyak1964some}, with step size and momentum computed at every step from the curvature of \(\Phi_k\) measured along the iterates. Verification compares the candidate with the anchor on the same \(\Phi_k\), with \(\boldsymbol{\mu}_k\) as fallback. The core interface is

\begin{equation}
    \begin{aligned}
    \text{problem:}\quad & \min_{\mathbf{z}}\ \Phi_k(\mathbf{z}),\\
    \text{proposal:}\quad & \mathbf{z}_k^{(J-1)}=\mathrm{HB}^{J-1}_{\Phi_k}(\boldsymbol{\mu}_k),\\
    \text{verification:}\quad & \mathbf{z}_k^{+}=\operatorname*{arg\,min}_{\mathbf{z}\in\{\boldsymbol{\mu}_k,\,\mathbf{z}_k^{(J-1)}\}}\Phi_k(\mathbf{z}),
    \end{aligned}
    \label{eq:propose_verify_principle}
\end{equation}
where \(\mathrm{HB}^{J-1}_{\Phi_k}(\boldsymbol{\mu}_k)\) denotes \(J-1\) heavy-ball steps on \(\Phi_k\) started at \(\boldsymbol{\mu}_k\) (Section~\ref{sec:method_proposal_verification}) and a tie keeps \(\boldsymbol{\mu}_k\). Both parts are derived from this problem. The proposal is the heavy-ball method with its step size and momentum matched to the curvature of \(\Phi_k\): the proximal term contributes exactly \(\beta_k\), the curvature \(w\) of the data term is measured along the iterates and carried from step to step, because every \(\Phi_k\) has the same data term, and Polyak's step size and momentum are optimal for quadratics with curvature in \([\beta_k,\beta_k+w]\)~\cite{polyak1964some}; the first step, which has no momentum, uses the step size \(1/M\), \(M=\beta_k+w\), the \(\alpha\) that maximizes the bound \(\alpha(1-\alpha M/2)\|\nabla\Phi_k\|^2\) on the decrease of \(\Phi_k\) when its curvature is at most \(M\) (both are derived in Appendix~\ref{app:polyak}). For a forward model that clips its output, the steps follow the gated derivative of Section~\ref{sec:method_objective_smoothing}, while the verifier keeps the exact objective. Verification follows from~\eqref{eq:trajectory_objective}: \(\mathbf{z}=\boldsymbol{\mu}_k\) adds no information, so a correction is committed only if it lowers its term of \(\mathcal{J}\) below that of \(\boldsymbol{\mu}_k\), whose value \(\Phi_k(\boldsymbol{\mu}_k)=\ell(F(\boldsymbol{\mu}_k),\mathbf{y})\) is known from the first gradient evaluation. Hence \(\Phi_k(\mathbf{z}_k^+)\leq\Phi_k(\boldsymbol{\mu}_k)\) whatever the proposal does.

This follows a decoupled-conditioning viewpoint, where measurement-consistency computation is separated from denoising or noise evolution, as in decomposed diffusion sampling and DAPS/LatentDAPS \cite{chung2024decomposed,zhang2025improving}. \textsc{PIVOT} instantiates this principle within a single reverse step: a measured-curvature inertial branch proposes a likelihood-informed clean correction, the frozen objective verifies it, and fresh noising returns the accepted clean state to the diffusion trajectory. This differs from raw likelihood guidance, where the measurement gradient is applied directly to the reverse update and the guidance scale controls both correction strength and risk. Although the proposal uses momentum~\cite{he2023fast,wizadwongsa2024diffusion}, inertia is confined to the local proposal, while acceptance is determined by verification.

Figure~\ref{fig:proposal_verification} summarizes the resulting propose--verify--reinsert step, while Algorithm~\ref{alg:pvppc} gives the complete procedure.

\subsection{Frozen objective}
\label{sec:method_objective_smoothing}

At step \(k\), \(\boldsymbol{\mu}_k\) is fixed while the correction is computed. We set \(q_k=\max(\sigma_k,\sigma_{\rm floor})\) and \(\beta_k=q_k^{-2}\), where \(\sigma_{\rm floor}>0\) avoids singular precision near zero noise. This corresponds to an isotropic local surrogate \(\mathbf{z}_0\mid\mathbf{u}_k\approx\mathcal{N}(\boldsymbol{\mu}_k,q_k^2I)\). The frozen objective used for verification is Eq.~\eqref{eq:frozen_objective}; ``frozen'' means that \(\boldsymbol{\mu}_k\), \(\beta_k\), and \(\mathbf{y}\) are fixed during the local correction.

\paragraph{Clipping forward models.} When a forward model clips its output (high dynamic range, and the nonlinear blur operator, whose output is clipped to the image range), a pixel predicted beyond the clip receives no data force even when its measurement shows that it is not saturated. In the gradients of \(\ell\), \textsc{PIVOT} therefore passes the gradient through such a pixel when the negative derivative of its misfit points back into range and its misfit exceeds the threshold \(\tau=\sigma_n\sqrt{2\ln N}\) (\(N\) measurements per image; the test concerns this pixel's own derivative, and through the Jacobian of the forward model, the proximal term and the momentum a step can still move the pixel either way); \(\Phi_k\), the restart test and the verifier keep the exact clip (Appendix~\ref{app:gate}). With this threshold, a pixel that is truly saturated can be released from its saturated side only if noise puts its measurement more than \(\tau\) inside the range, so the expected number of such pixels per image is at most \(1/\sqrt{4\pi\ln N}\) (Proposition~\ref{prop:gate_release}).

For latent diffusion, \(\mathbf{u}_k\), \(\boldsymbol{\mu}_k\), and \(\mathbf{z}\) all live in latent space. We set \(G=\mathrm{Dec}\), so the measurement loss is \(\ell(A(\mathrm{Dec}(\mathbf{z})),\mathbf{y})\), and proposal gradients backpropagate through both the decoder and measurement operator. Fresh-noise reinsertion is also applied in latent space. The final reconstruction is decoded as \(\hat{\mathbf{x}}=\mathrm{Dec}(\mathbf{u}_K)\).

\subsection{Heavy-ball proposals and verified selection}
\label{sec:method_proposal_verification}

At each corrected reverse step, initialize \(\mathbf{z}_k^{(0)}=\boldsymbol{\mu}_k\) and \(\mathbf{z}_k^{(-1)}=\mathbf{z}_k^{(0)}\). For \(j=0,\ldots,J-2\), define

\[
    \mathbf{g}_k^{(j)}
    =
    \nabla_{\mathbf{z}}\ell(F(\mathbf{z}_k^{(j)}),\mathbf{y})
    +
    \beta_k(\mathbf{z}_k^{(j)}-\boldsymbol{\mu}_k).
\]
Here \(\nabla_{\mathbf{z}}\ell\) is the gradient returned by automatic differentiation, with the gated clip derivative of Section~\ref{sec:method_objective_smoothing} for forward models that clip their output (Appendix~\ref{app:theory_setting}). The heavy-ball update is
\begin{equation}
    \mathbf{z}_k^{(j+1)}=\mathbf{z}_k^{(j)}-\alpha\,\mathbf{g}_k^{(j)}+\gamma\,(\mathbf{z}_k^{(j)}-\mathbf{z}_k^{(j-1)}),\qquad
    \alpha=\frac{4}{(\sqrt{M}+\sqrt{m})^2},\quad \gamma=\Big(\frac{\sqrt{M}-\sqrt{m}}{\sqrt{M}+\sqrt{m}}\Big)^2,
    \label{eq:hb_update}
\end{equation}
Polyak's optimal step size and momentum for a quadratic whose curvature lies in \([m,M]\) \cite{polyak1964some}. The proximal term contributes curvature exactly \(\beta_k\), which is taken as the lower end \(m=\beta_k\) (the data term is modelled with curvature in \([0,w]\)). The upper end is \(M=w+\beta_k\), where \(w\) estimates the curvature of the data term by the largest guarded secant \(\|\nabla\ell(F(\mathbf{z}'),\mathbf{y})-\nabla\ell(F(\mathbf{z}),\mathbf{y})\|/\|\mathbf{z}'-\mathbf{z}\|\) observed along the iterates, a gradient-secant estimate in the spirit of Barzilai and Borwein~\cite{barzilai1988two}; the value with which step \(k+1\) starts is carried from step \(k\) with a decay, \(w_{k+1}=\max(V_k,w_k/2)\), where \(w_k\) is the value with which step \(k\) started and \(V_k\) is the largest guarded secant observed at step \(k\). At \(j=0\) there is no momentum, and the formula is evaluated with \(m=M\), which gives \(\alpha=1/M\) and \(\gamma=0\). 
\begin{proposition}[Proposal parameters]\label{prop:proposal_params}
Let \(\Phi\) be a quadratic whose Hessian has its eigenvalues in \([m,M]\), \(0<m\le M\).
(i) A gradient step of size \(\alpha\) lowers \(\Phi\) by at least \(\alpha(1-\alpha M/2)\|\nabla\Phi\|_2^2\), and this bound is largest at \(\alpha=1/M\).
(ii) With \(\alpha\) and \(\gamma\) of Eq.~\eqref{eq:hb_update}, every eigen-direction of the heavy-ball iteration has the worst-case asymptotic rate \(q^\star=(\sqrt{M/m}-1)/(\sqrt{M/m}+1)\), and no \(\alpha>0\) and \(\gamma\in[0,1)\) give a smaller rate at both \(m\) and \(M\).
\end{proposition}
PIVOT applies Proposition~\ref{prop:proposal_params} with \(m=\beta_k\) and \(M=\beta_k+w\). For any forward model, \([\beta_k,\beta_k+w]\) is only a working curvature model: \(w\) is estimated from gradient secants and need not bound the curvature of the data term, which for a nonlinear \(F\) (a nonlinear operator or a latent decoder) can also be negative, so that \(\beta_k\) need not be a lower bound either. The proposition therefore motivates the parameters (proof in Appendix~\ref{app:polyak}), and the certificate of Proposition~\ref{prop:pvppc_scale_displacement} holds whatever this model is. If a step increases \(\Phi_k\), the point is kept and the momentum is dropped, as in function-value restart of accelerated methods~\cite{odonoghue2015adaptive}; in addition, \(w\) is doubled. Thus \(\alpha\) and \(\gamma\) are recomputed at every step from the measured curvature and are not tuned.
The loop is a candidate generator, not an inner sampler or exact optimizer.

The verifier compares the final iterate with the anchor and commits the better one, the minimizer in the verification line of \eqref{eq:propose_verify_principle},
\begin{equation}
    \mathbf{z}_k^+=\begin{cases}\mathbf{z}_k^{(J-1)}&\text{if }\Phi_k(\mathbf{z}_k^{(J-1)})<\Phi_k(\boldsymbol{\mu}_k),\\ \boldsymbol{\mu}_k&\text{otherwise.}\end{cases}
    \label{eq:best_candidate_verification}
\end{equation}
Because \(\boldsymbol{\mu}_k\) is one of the two candidates, exact verification gives \(\Phi_k(\mathbf{z}_k^+)\leq \Phi_k(\boldsymbol{\mu}_k)\). 

 The guarantee is local to the frozen objective only; it is not a global convergence theorem, exact posterior-sampling guarantee, or calibrated uncertainty statement. After verification, fresh noising reinserts the accepted clean state:
\begin{equation}
    \mathbf{u}_{k+1}
    =
    \mathbf{z}_k^+
    +
    \sigma_{k+1}\boldsymbol{\epsilon}_k,
    \qquad
    \boldsymbol{\epsilon}_k\sim\mathcal{N}(\mathbf{0},I).
    \label{eq:fresh_transition}
\end{equation}
The final reconstruction is \(\hat{\mathbf{x}}=G(\mathbf{u}_K)\)\({}=G(\mathbf z^+_{K-1})\), since the schedule ends at \(\sigma_K=0\) (Appendix~\ref{app:theory_setting}).

\subsection{Local theory and implications}
\label{sec:method_theory}

We summarize the per-step consequences that guide the design; assumptions and proofs are in Appendix~\ref{app:theory}. Let
\(L(\mathbf z)=\ell(F(\mathbf z),\mathbf y)\) and
\(\Phi(\mathbf z)=L(\mathbf z)+\frac{\beta}{2}\|\mathbf z-\boldsymbol{\mu}\|^2\).

\begin{proposition}[Per-step certificate]
\label{prop:pvppc_scale_displacement}

Whatever produced the candidate, exact verification gives \(\Phi(\mathbf z^+)=\min\{\Phi(\boldsymbol{\mu}),\Phi(\mathbf z^{(J-1)})\}\leq\Phi(\boldsymbol{\mu})=L(\boldsymbol{\mu})\).
With \(\boldsymbol{\Delta}^+=\mathbf z^+-\boldsymbol{\mu}\),
\[
    \frac{\beta}{2}\|\boldsymbol{\Delta}^+\|_2^2
    \leq
    L(\boldsymbol{\mu})-L(\mathbf z^+).
\]
For fresh-noise reinsertion with \(\sigma_{k+1}>0\), \(Q^+=\mathcal N(\mathbf z^+,\sigma_{k+1}^2I)\) and \(Q^0=\mathcal N(\boldsymbol{\mu},\sigma_{k+1}^2I)\) satisfy
\[
    \mathrm{KL}(Q^+\|Q^0)
    \leq
    \frac{L(\boldsymbol{\mu})-L(\mathbf z^+)}
    {\beta\sigma_{k+1}^2}.
\]
\end{proposition}

Proposition~\ref{prop:pvppc_scale_displacement} says that any accepted displacement must be paid for by data-loss improvement, controlling what is reinserted into the reverse trajectory{}, whatever the proposal.

\subsection{Computational cost}
\label{sec:method_cost}

Each outer step uses one denoiser evaluation to obtain \(\boldsymbol{\mu}_k\).  At each step, the proposal uses \(J\) data-gradient evaluations, each one forward/backward pass through \(F=A\circ G\) (plus one probe gradient per run, at the first step, to initialize the curvature estimate; Algorithm~\ref{alg:pvppc}). Verification adds only one gradient-free evaluation of \(\Phi_k\), at the final iterate, since \(\Phi_k(\boldsymbol{\mu}_k)=L(\boldsymbol{\mu}_k)\) is known from the first gradient evaluation; the restart tests reuse the objective values computed with the gradients. Thus the per-step cost is controlled by \(J\), with no additional denoiser calls inside proposal or verification. In latent diffusion, each evaluation of \(F\) also includes the decoder.
\section{Experiments}
\label{sec:Experiments}
\begin{table}
\caption{Quantitative evaluation on FFHQ and ImageNet using RTX 6000 Ada. The best and second-best results within each task type (Pixel and Latent) are indicated in \textbf{bold} and \underline{underlined}, respectively.}
\label{tab:main_rtx6000ada}
\centering
\begin{adjustbox}{width=0.99\textwidth}
\begin{tabular}{l!{\vrule}l!{\vrule}l!{\vrule}cccc!{\vrule}cccc}
\toprule
\multirow{2}{*}{\textbf{Task}} & \multirow{2}{*}{\textbf{Type}} & \multirow{2}{*}{\textbf{Method}} & \multicolumn{4}{c|}{\textbf{FFHQ}} & \multicolumn{4}{c}{\textbf{ImageNet}} \\
 & & & \textbf{PSNR ($\uparrow$)} & \textbf{SSIM ($\uparrow$)} & \textbf{LPIPS ($\downarrow$)} & \textbf{Runtime (s) ($\downarrow$)} & \textbf{PSNR ($\uparrow$)} & \textbf{SSIM ($\uparrow$)} & \textbf{LPIPS ($\downarrow$)} & \textbf{Runtime (s) ($\downarrow$)} \\
\midrule

\multirow{11}{*}{Super resolution 4$\times$} & \multirow{6}{*}{\textit{Pixel}} & Ours & \textbf{29.666} & \textbf{0.847} & 0.249 & \textbf{1.305} & \textbf{26.618} & \textbf{0.731} & \underline{0.325} & \textbf{2.398} \\
 &  & FAST-DIPS & \underline{29.573} & \underline{0.841} & 0.244 & \underline{1.394} & 26.367 & 0.714 & 0.334 & \underline{3.987} \\
 &  & DAPS & 28.619 & 0.764 & 0.262 & 28.257 & 25.512 & 0.636 & 0.374 & 65.580 \\
 &  & SITCOM & 29.153 & 0.826 & \underline{0.231} & 15.755 & \underline{26.519} & \underline{0.716} & \textbf{0.309} & 60.778 \\
 &  & DMPlug & 28.637 & 0.797 & 0.253 & 118.255 & 25.091 & 0.660 & 0.347 & 238.433 \\
 &  & DCDP & 27.611 & 0.785 & \textbf{0.225} & 4.719 & 24.358 & 0.640 & 0.369 & 9.190 \\
\cmidrule{2-11}
 & \multirow{5}{*}{\textit{Latent}} & Ours & \underline{29.033} & \underline{0.824} & 0.292 & \textbf{39.816} & 25.998 & 0.695 & 0.402 & \textbf{39.112} \\
 &  & Latent FAST-DIPS & 28.634 & 0.797 & \underline{0.283} & \underline{42.537} & \underline{26.005} & \underline{0.697} & \textbf{0.371} & \underline{40.512} \\
 &  & Latent DAPS & \textbf{29.204} & \textbf{0.825} & \textbf{0.272} & 84.460 & \textbf{26.189} & \textbf{0.702} & \underline{0.388} & 86.315 \\
 &  & ReSample & 23.317 & 0.456 & 0.507 & 300.061 & 22.152 & 0.423 & 0.470 & 269.078 \\
 &  & PSLD & 24.627 & 0.628 & 0.377 & 69.290 & 16.656 & 0.291 & 0.541 & 102.757 \\
\midrule

\multirow{11}{*}{Inpaint (box)} & \multirow{6}{*}{\textit{Pixel}} & Ours & \underline{25.131} & \textbf{0.860} & \textbf{0.156} & \underline{2.403} & \textbf{21.705} & \textbf{0.803} & \textbf{0.230} & \underline{5.159} \\
 &  & FAST-DIPS & 24.605 & \underline{0.850} & 0.190 & \textbf{1.230} & \underline{21.381} & \underline{0.777} & 0.278 & \textbf{3.958} \\
 &  & DAPS & 24.285 & 0.746 & 0.222 & 22.352 & 21.234 & 0.720 & 0.270 & 57.108 \\
 &  & SITCOM & \textbf{25.275} & 0.831 & \underline{0.179} & 17.697 & 20.698 & 0.726 & \underline{0.241} & 66.709 \\
 &  & DMPlug & 23.526 & 0.790 & 0.281 & 103.760 & 19.455 & 0.667 & 0.384 & 253.946 \\
 &  & DCDP & 21.721 & 0.766 & 0.219 & 8.147 & 17.741 & 0.662 & 0.301 & 17.214 \\
\cmidrule{2-11}
 & \multirow{5}{*}{\textit{Latent}} & Ours & \textbf{24.584} & \textbf{0.838} & \textbf{0.243} & 80.738 & \underline{21.229} & \textbf{0.751} & \underline{0.341} & \underline{78.573} \\
 &  & Latent FAST-DIPS & \underline{24.084} & \underline{0.829} & \underline{0.247} & \textbf{40.586} & 19.576 & \underline{0.715} & 0.379 & \textbf{42.472} \\
 &  & Latent DAPS & 23.474 & 0.742 & 0.369 & 85.823 & \textbf{22.818} & 0.561 & 0.543 & 88.725 \\
 &  & ReSample & 19.978 & 0.796 & \underline{0.247} & 296.326 & 18.087 & 0.713 & \textbf{0.309} & 265.139 \\
 &  & PSLD & 21.847 & 0.612 & 0.370 & \underline{69.045} & 18.762 & 0.424 & 0.549 & 99.832 \\
\midrule

\multirow{11}{*}{Inpaint (random)} & \multirow{6}{*}{\textit{Pixel}} & Ours & \textbf{32.641} & \textbf{0.914} & \textbf{0.144} & \underline{2.644} & \textbf{29.975} & \textbf{0.858} & \textbf{0.181} & \underline{5.738} \\
 &  & FAST-DIPS & 31.022 & \underline{0.879} & 0.202 & \textbf{1.233} & 28.353 & 0.791 & 0.249 & \textbf{3.929} \\
 &  & DAPS & 30.078 & 0.791 & 0.214 & 21.576 & 27.350 & 0.721 & 0.251 & 56.932 \\
 &  & SITCOM & \underline{31.431} & 0.871 & \underline{0.173} & 25.382 & \underline{29.008} & \underline{0.821} & \underline{0.183} & 95.962 \\
 &  & DMPlug & 30.825 & 0.865 & 0.231 & 122.143 & 27.023 & 0.739 & 0.317 & 248.146 \\
 &  & DCDP & 27.240 & 0.788 & 0.229 & 8.130 & 23.284 & 0.606 & 0.355 & 17.518 \\
\cmidrule{2-11}
 & \multirow{5}{*}{\textit{Latent}} & Ours & \textbf{31.644} & \textbf{0.890} & \underline{0.210} & \textbf{32.613} & \textbf{28.327} & \textbf{0.801} & \underline{0.282} & \textbf{33.775} \\
 &  & Latent FAST-DIPS & \underline{30.091} & \underline{0.877} & \textbf{0.201} & \underline{39.644} & \underline{26.925} & \underline{0.769} & 0.291 & \underline{40.436} \\
 &  & Latent DAPS & 26.036 & 0.743 & 0.385 & 85.840 & 19.630 & 0.588 & 0.522 & 86.092 \\
 &  & ReSample & 29.950 & 0.842 & \textbf{0.201} & 307.874 & 26.916 & 0.756 & \textbf{0.255} & 323.283 \\
 &  & PSLD & 24.280 & 0.635 & 0.346 & 68.915 & 20.690 & 0.436 & 0.550 & 98.449 \\
\midrule

\multirow{11}{*}{Gaussian deblurring} & \multirow{6}{*}{\textit{Pixel}} & Ours & \textbf{29.630} & \textbf{0.842} & \underline{0.239} & \underline{3.675} & \underline{26.516} & \underline{0.715} & \underline{0.338} & \underline{4.540} \\
 &  & FAST-DIPS & \underline{29.406} & \underline{0.836} & 0.247 & \textbf{1.563} & 26.181 & 0.705 & 0.344 & \textbf{3.461} \\
 &  & DAPS & 28.762 & 0.767 & 0.254 & 37.730 & 25.932 & 0.655 & 0.355 & 74.984 \\
 &  & SITCOM & 29.359 & 0.826 & \textbf{0.236} & 25.491 & \textbf{26.741} & \textbf{0.719} & \textbf{0.315} & 95.161 \\
 &  & DMPlug & 27.950 & 0.770 & 0.278 & 181.695 & 24.082 & 0.613 & 0.389 & 313.680 \\
 &  & DCDP & 27.553 & 0.761 & 0.245 & 4.934 & 23.499 & 0.542 & 0.449 & 9.573 \\
\cmidrule{2-11}
 & \multirow{5}{*}{\textit{Latent}} & Ours & \textbf{28.047} & \textbf{0.796} & \underline{0.325} & \underline{50.341} & \underline{24.965} & \underline{0.644} & 0.446 & \underline{50.784} \\
 &  & Latent FAST-DIPS & \underline{28.006} & \underline{0.793} & \textbf{0.312} & \textbf{42.393} & \textbf{25.283} & \textbf{0.658} & \textbf{0.427} & \textbf{42.260} \\
 &  & Latent DAPS & 25.717 & 0.732 & 0.384 & 87.407 & 22.695 & 0.567 & 0.549 & 86.016 \\
 &  & ReSample & 26.357 & 0.662 & 0.329 & 355.885 & 23.530 & 0.497 & \underline{0.439} & 338.281 \\
 &  & PSLD & 22.015 & 0.503 & 0.444 & 70.098 & 19.591 & 0.329 & 0.555 & 109.263 \\
\midrule

\multirow{11}{*}{Motion deblurring} & \multirow{6}{*}{\textit{Pixel}} & Ours & \textbf{32.413} & \textbf{0.894} & 0.173 & \underline{1.627} & \textbf{29.849} & \textbf{0.834} & \textbf{0.212} & \textbf{3.282} \\
 &  & FAST-DIPS & \underline{31.736} & \underline{0.878} & \underline{0.171} & \textbf{1.574} & 29.037 & 0.799 & 0.236 & \underline{3.516} \\
 &  & DAPS & 30.954 & 0.823 & 0.202 & 38.606 & 28.623 & 0.764 & 0.246 & 78.000 \\
 &  & SITCOM & 31.501 & 0.866 & \textbf{0.167} & 26.184 & \underline{29.321} & \underline{0.811} & \underline{0.213} & 95.529 \\
 &  & DMPlug & 29.272 & 0.818 & 0.260 & 152.699 & 25.211 & 0.660 & 0.365 & 284.413 \\
 &  & DCDP & 25.468 & 0.566 & 0.365 & 5.007 & 17.771 & 0.248 & 0.574 & 9.488 \\
\cmidrule{2-11}
 & \multirow{5}{*}{\textit{Latent}} & Ours & \textbf{29.936} & \textbf{0.838} & \underline{0.278} & \underline{50.678} & \textbf{26.791} & \textbf{0.717} & \textbf{0.378} & \underline{54.401} \\
 &  & Latent FAST-DIPS & \underline{29.285} & \underline{0.822} & \underline{0.278} & \textbf{42.419} & \underline{26.623} & \underline{0.708} & \underline{0.387} & \textbf{43.119} \\
 &  & Latent DAPS & 26.646 & 0.757 & 0.361 & 87.409 & 23.557 & 0.592 & 0.513 & 89.050 \\
 &  & ReSample & 28.744 & 0.754 & \textbf{0.262} & 347.756 & 24.845 & 0.579 & 0.404 & 311.967 \\
 &  & PSLD & 24.416 & 0.603 & 0.346 & 70.041 & 20.761 & 0.376 & 0.518 & 98.400 \\
\midrule

\multirow{9}{*}{Phase retrieval} & \multirow{5}{*}{\textit{Pixel}} & Ours & \textbf{31.343} & \textbf{0.880} & \textbf{0.205} & \underline{24.051} & \underline{20.958} & \textbf{0.547} & 0.435 & \underline{28.194} \\
 &  & FAST-DIPS & 29.253 & \underline{0.851} & 0.218 & \textbf{5.220} & 20.945 & \underline{0.533} & \underline{0.434} & \textbf{10.217} \\
 &  & DAPS & \underline{30.103} & 0.796 & \underline{0.208} & 98.054 & \textbf{22.354} & 0.519 & \textbf{0.402} & 241.113 \\
 &  & SITCOM & 28.782 & 0.791 & 0.239 & 24.584 & 18.481 & 0.383 & 0.524 & 94.187 \\
 &  & DCDP & 24.036 & 0.677 & 0.310 & 102.386 & 15.953 & 0.283 & 0.595 & 195.193 \\
\cmidrule{2-11}
 & \multirow{4}{*}{\textit{Latent}} & Ours & \underline{27.925} & \underline{0.788} & \underline{0.304} & \underline{79.183} & \underline{18.166} & \underline{0.444} & \underline{0.589} & \underline{82.102} \\
 &  & Latent FAST-DIPS & \textbf{28.330} & \textbf{0.789} & \textbf{0.244} & \textbf{76.893} & \textbf{18.874} & 0.441 & \textbf{0.507} & \textbf{75.505} \\
 &  & Latent DAPS & 23.199 & 0.692 & 0.421 & 177.730 & 17.067 & \textbf{0.446} & 0.624 & 192.787 \\
 &  & ReSample & 24.676 & 0.606 & 0.412 & 320.911 & 16.913 & 0.320 & 0.608 & 319.601 \\
\midrule

\multirow{10}{*}{Nonlinear deblurring} & \multirow{6}{*}{\textit{Pixel}} & Ours & \textbf{29.939} & \textbf{0.848} & 0.221 & 151.812 & \textbf{28.562} & \textbf{0.795} & \underline{0.259} & 157.840 \\
 &  & FAST-DIPS & 27.818 & 0.803 & 0.280 & \textbf{24.587} & 25.607 & 0.695 & 0.373 & \textbf{30.277} \\
 &  & DAPS & 28.868 & 0.780 & 0.223 & 755.725 & 27.537 & 0.734 & 0.266 & 884.119 \\
 &  & SITCOM & \underline{29.519} & \underline{0.812} & \underline{0.207} & \underline{27.167} & \underline{28.097} & \underline{0.774} & \textbf{0.221} & \underline{97.029} \\
 &  & DMPlug & 28.298 & 0.811 & 0.249 & 291.221 & 25.086 & 0.679 & 0.317 & 668.641 \\
 &  & DCDP & 27.879 & 0.795 & \textbf{0.204} & 269.786 & 25.726 & 0.664 & 0.280 & 367.695 \\
\cmidrule{2-11}
 & \multirow{4}{*}{\textit{Latent}} & Ours & \textbf{29.131} & \textbf{0.824} & 0.287 & 200.075 & \textbf{26.279} & \textbf{0.710} & \underline{0.379} & 213.401 \\
 &  & Latent FAST-DIPS & 28.746 & \underline{0.823} & \underline{0.260} & \textbf{86.419} & 24.370 & 0.633 & 0.465 & \textbf{84.354} \\
 &  & Latent DAPS & 25.152 & 0.726 & 0.387 & \underline{194.014} & 22.516 & 0.568 & 0.530 & \underline{198.074} \\
 &  & ReSample & \underline{28.748} & 0.797 & \textbf{0.236} & 843.212 & \underline{26.047} & \underline{0.697} & \textbf{0.301} & 686.128 \\
\midrule

\multirow{9}{*}{High dynamic range} & \multirow{5}{*}{\textit{Pixel}} & Ours & 26.013 & \textbf{0.859} & \textbf{0.186} & \textbf{1.537} & 25.379 & \textbf{0.848} & \textbf{0.190} & \textbf{2.548} \\
 &  & FAST-DIPS & 26.275 & \underline{0.843} & 0.218 & \underline{3.166} & 24.522 & 0.775 & 0.290 & \underline{8.424} \\
 &  & DAPS & \underline{27.149} & 0.834 & \underline{0.196} & 76.724 & \textbf{26.568} & \underline{0.819} & \underline{0.198} & 216.701 \\
 &  & SITCOM & \textbf{27.205} & 0.777 & 0.226 & 33.887 & \underline{26.449} & 0.774 & 0.222 & 129.681 \\
 &  & DMPlug & 25.507 & 0.777 & 0.264 & 221.418 & 23.244 & 0.715 & 0.308 & 564.501 \\
\cmidrule{2-11}
 & \multirow{4}{*}{\textit{Latent}} & Ours & \underline{25.050} & \underline{0.826} & 0.260 & \textbf{63.385} & \textbf{25.547} & \underline{0.779} & \underline{0.292} & \textbf{65.486} \\
 &  & Latent FAST-DIPS & \textbf{25.869} & \textbf{0.832} & \underline{0.247} & \underline{78.769} & 24.026 & 0.766 & 0.301 & \underline{74.249} \\
 &  & Latent DAPS & 20.789 & 0.630 & 0.512 & 174.976 & 19.394 & 0.469 & 0.641 & 180.929 \\
 &  & ReSample & 25.038 & 0.822 & \textbf{0.239} & 291.372 & \underline{24.950} & \textbf{0.783} & \textbf{0.257} & 273.695 \\
\bottomrule
\end{tabular}
\end{adjustbox}
\end{table}
\subsection{Experimental setup}
\label{subsec:Experimental setup}

For natural image inverse problems, we follow the experimental protocol of DAPS~\citep{zhang2025improving}. We evaluate on five linear and three nonlinear inverse problems using FFHQ $256{\times}256$~\citep{karras2019style} and ImageNet $256{\times}256$~\citep{deng2009imagenet}, with 100 evaluation images from each dataset. For pixel-space experiments, all methods use the same pretrained diffusion checkpoints: the FFHQ checkpoint from~\citep{chung2023diffusion} and the ImageNet checkpoint from~\citep{dhariwal2021diffusion}. For latent-space experiments, all methods use the same LDM-VQ4 checkpoint for each dataset, an unconditional FFHQ model and a class-conditional ImageNet model~\citep{rombach2022high}. Pixel-space baselines include FAST-DIPS~\citep{kim2026fastdips}, DAPS~\citep{zhang2025improving}, SITCOM~\citep{alkhouri2025sitcom}, DMPlug~\citep{wang2024dmplug}, and DCDP~\citep{li2024decoupled}. Latent-space baselines include Latent FAST-DIPS~\citep{kim2026fastdips}, Latent DAPS~\citep{zhang2025improving}, PSLD~\citep{rout2023solving}, and ReSample~\citep{song2024solving}.

For medical image reconstruction, we consider accelerated multi-coil knee MRI reconstruction on fastMRI~\citep{zbontar2018fastmri}, following the experimental protocol of InverseBench~\citep{zheng2025inversebench}. All MRI methods use the same pretrained diffusion checkpoint provided by InverseBench and are evaluated on the same test set of 96 samples. We simulate undersampled k-space measurements using random masks at acceleration factors of $\times4$ and $\times8$. MRI baselines include DPS~\citep{chung2023diffusion}, DAPS~\citep{zhang2025improving}, DDS~\citep{chung2024decomposed}, and FAST-DIPS~\citep{kim2026fastdips}.

For natural image experiments, we report PSNR, SSIM~\citep{wang2004ssim}, and LPIPS~\citep{Zhang_2018_CVPR}. For phase retrieval, we follow the best-of-four evaluation protocol used in DPS~\citep{chung2023diffusion}, selecting the run with the highest PSNR against ground truth (oracle selection) and computing SSIM and LPIPS on the same run. For all other natural image tasks, we report results from a single run per sample. For MRI reconstruction, we report PSNR and SSIM computed on magnitude images. All experiments are conducted on a single RTX 6000 Ada GPU. Since we use a batch size of one throughout, runtime is measured as the per-sample wall-clock sampling time: we record the pure sampling time for each individual sample and report the average over all samples. Additional experimental details are provided in Appendix~\ref{app:exp_detail}.

\subsection{Main results}
As shown in Table~\ref{tab:main_rtx6000ada}, \textsc{PIVOT} is broadly competitive with existing diffusion-based inverse solvers across FFHQ and ImageNet: it has the highest PSNR of its block in 11 of 16 pixel-space and 9 of 16 latent-space settings. Its strongest gains appear in random inpainting in both spaces and FFHQ pixel-space phase retrieval. The results are not uniformly dominant: FAST-DIPS, SITCOM, DAPS, or ReSample obtain better PSNR, SSIM, or LPIPS in some settings, including parts of super resolution, deblurring, box inpainting, and phase retrieval, and three of the four high dynamic range settings, where \textsc{PIVOT} trails the best baseline by 0.8--1.2 dB in PSNR. Thus the main empirical claim is robust performance across a broad set of inverse problems rather than universal superiority. Mechanistically, the gains come from the proposal: measured-curvature heavy-ball dynamics generate stronger local data-consistency candidates, while objective verification filters candidates that would worsen the frozen objective before they are reinserted into the reverse diffusion trajectory; in the component studies no final candidate was rejected (Section~\ref{subsec:ablation}). For linear inverse problems, the advantage is particularly apparent in random inpainting, where observed pixels can be corrected efficiently while missing regions remain governed by the diffusion prior. For nonlinear measurements, nonlinear deblurring shows strong PSNR/SSIM performance in pixel space, while on HDR \textsc{PIVOT} has the highest PSNR only in latent space on ImageNet. In latent space, the method remains competitive across the evaluated tasks.

\begin{figure*}[t] 
\centering
\vspace{-0.7em}
\setlength\tabcolsep{1pt} 
\resizebox{\textwidth}{!}{ 
\begin{tabular}{cccccc} 
\includegraphics[scale=0.095]{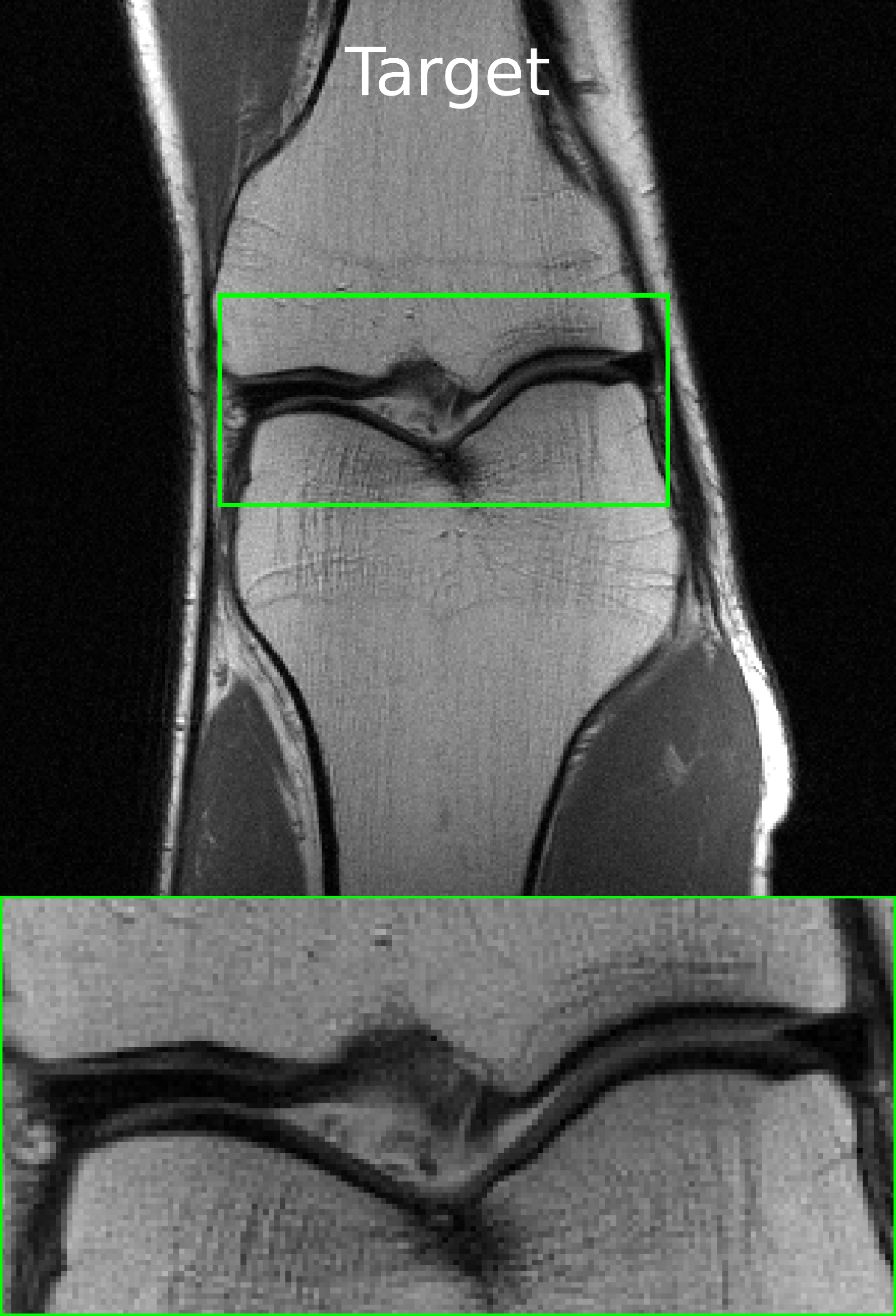} & 
\includegraphics[scale=0.095]{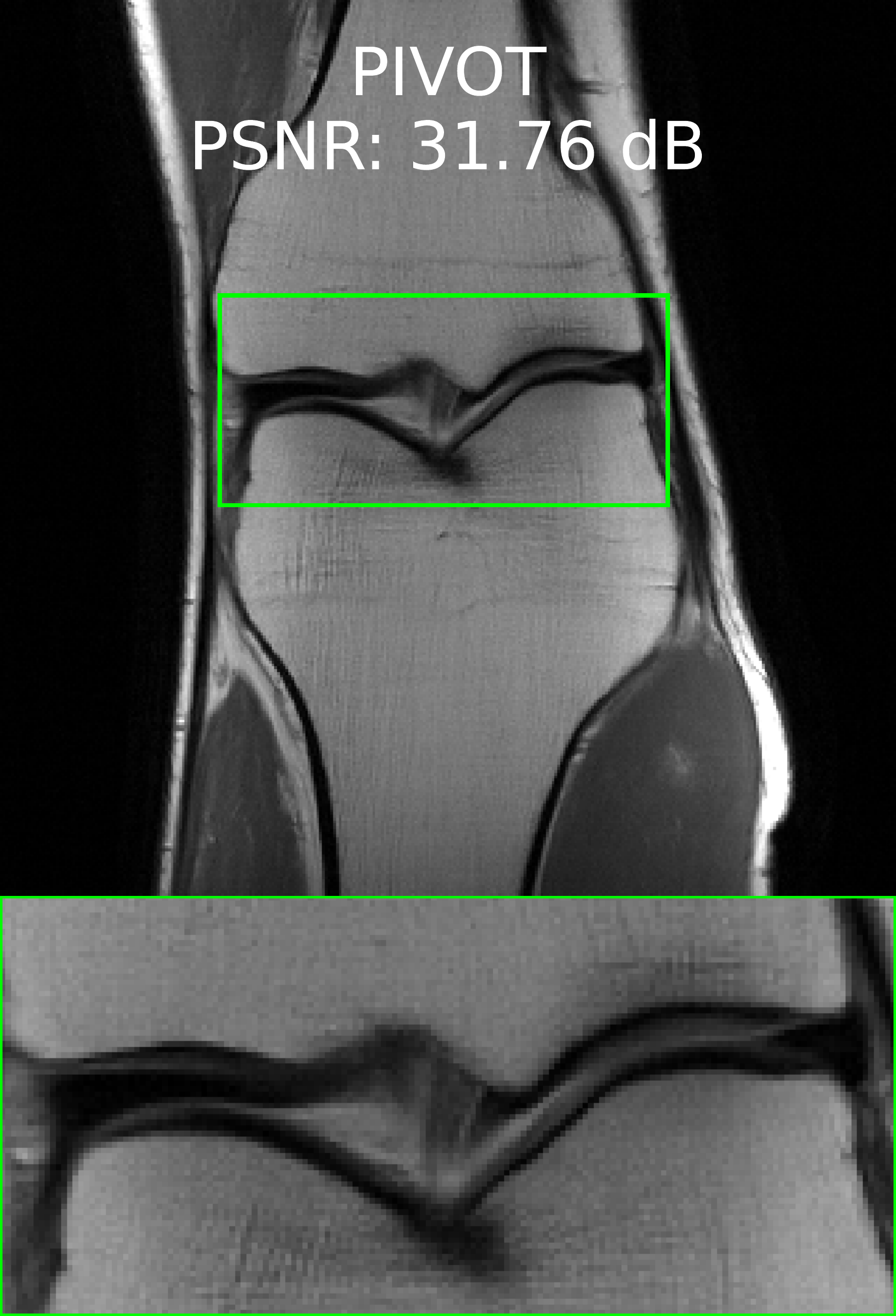} & 
\includegraphics[scale=0.095]{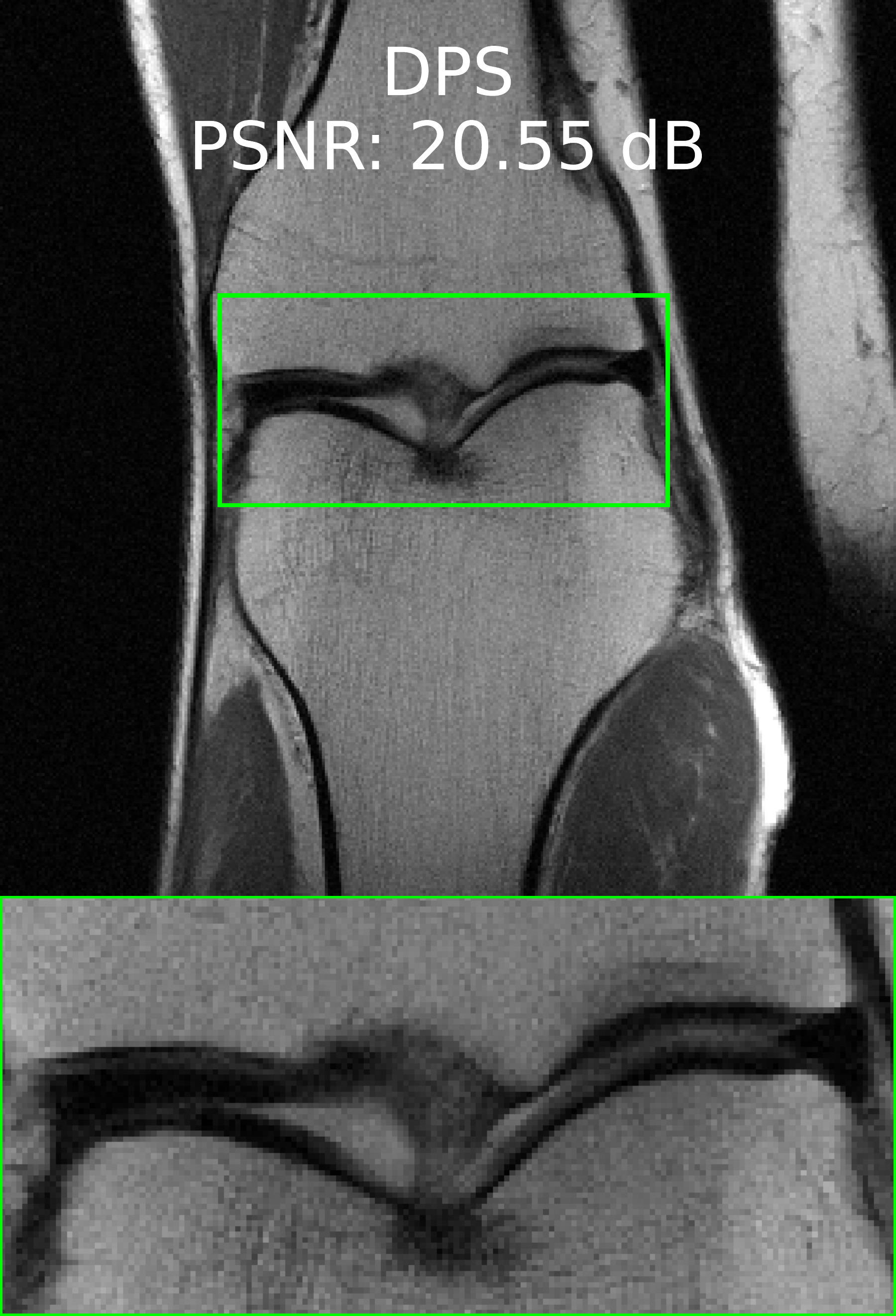} & 
\includegraphics[scale=0.095]{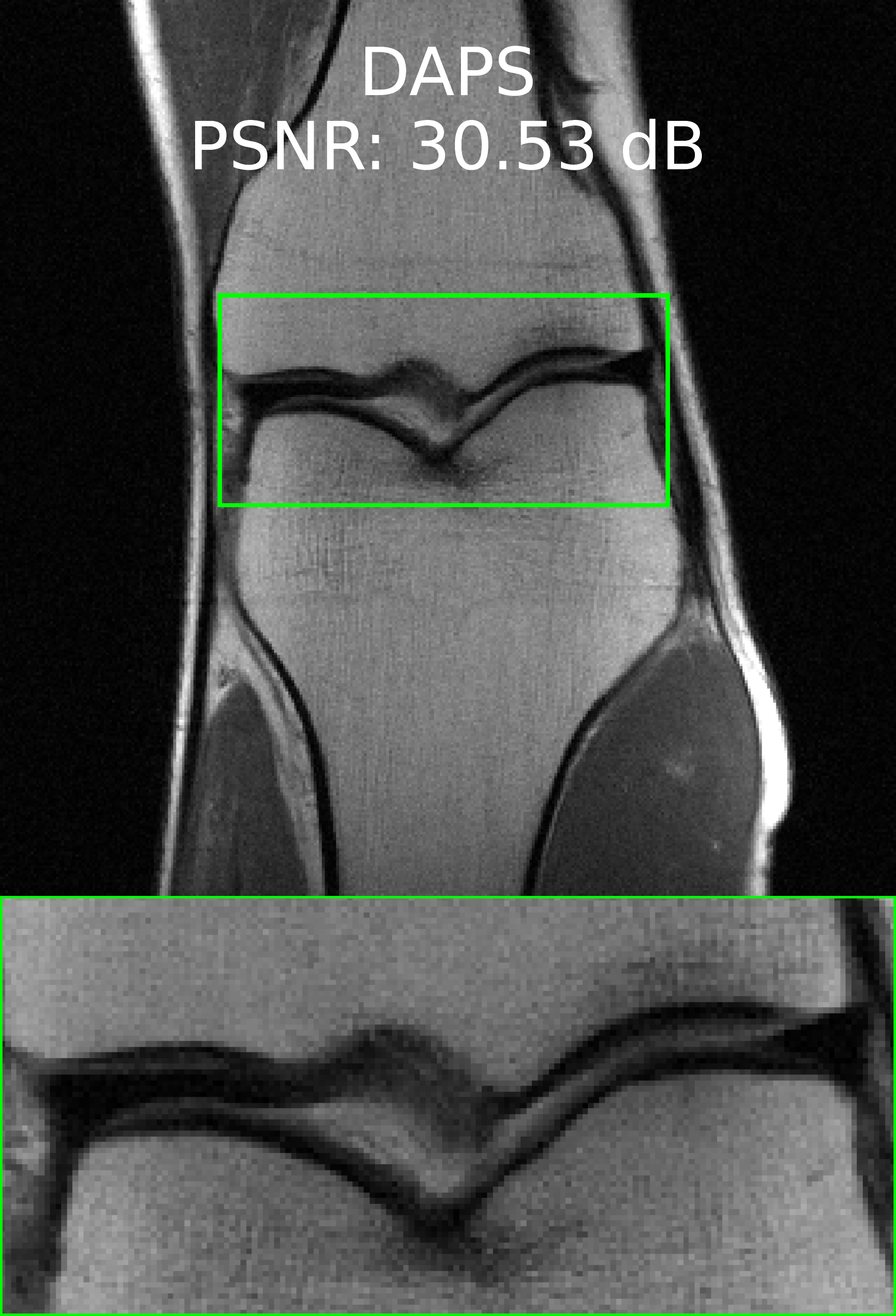} & 
\includegraphics[scale=0.095]{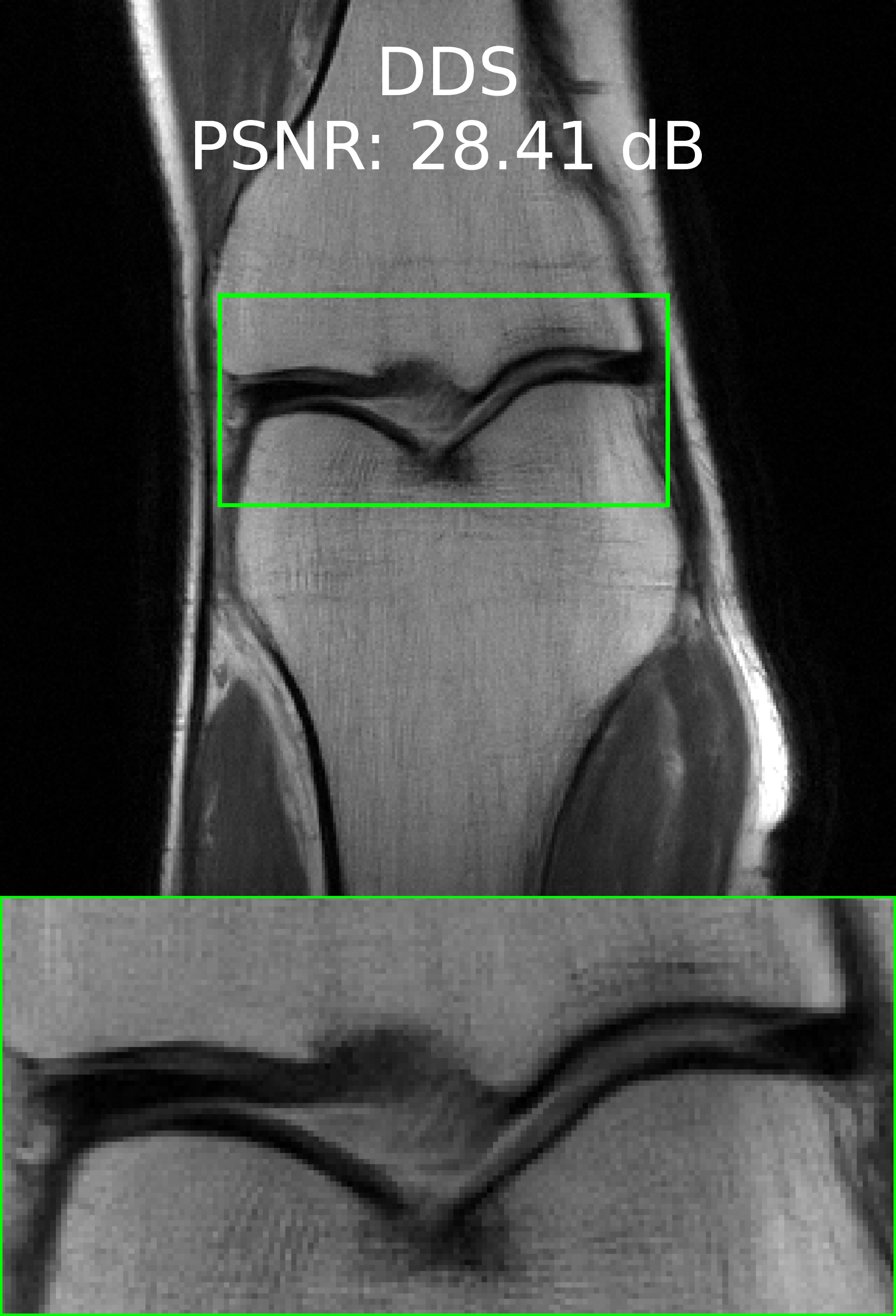} & 
\includegraphics[scale=0.095]{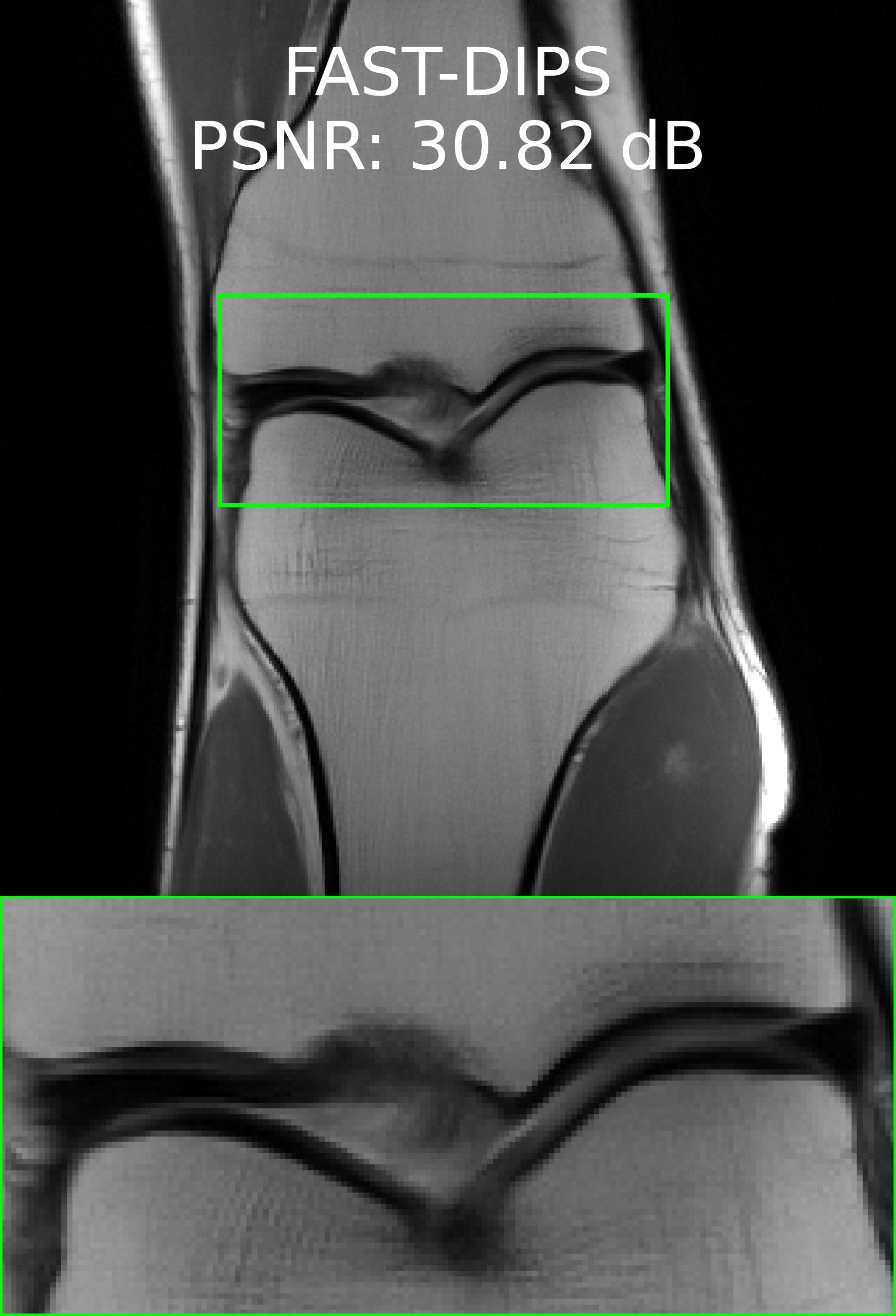} \\ 
\end{tabular}} 
\caption{Qualitative comparison for $\times8$ accelerated multi-coil fastMRI knee reconstruction. Each column shows the target image and the reconstruction from a different diffusion based method, with the corresponding PSNR reported in the image.} 
\vspace{-0.5em}
\label{fig:qual_mri}
\end{figure*}

Table~\ref{tab:mri_table} reports results on accelerated multi-coil knee MRI reconstruction under random-disc undersampling at acceleration factors of \(\times4\) and \(\times8\).  For SSIM, DAPS obtains the best value at both acceleration factors. Figure~\ref{fig:qual_mri} provides a qualitative comparison for the $\times8$ setting. The zoomed-in crops show local anatomical structures in the highlighted knee joint region.  Additional experimental results and analyses are provided in Appendix~\ref{app:add_exp}.

\begin{table}
\centering
\small
\caption{Quantitative evaluation for MRI reconstruction under random-disc undersampling at acceleration factors $\times4$ and $\times8$. We report average PSNR/SSIM over 96 test samples. Best scores are in \textbf{bold} and second-best are \underline{underlined}.}
\label{tab:mri_table}
\begin{tabular}{c c c c c}
\toprule
\multirow{2}{*}{\textbf{Method}} & \multicolumn{2}{c}{\textbf{$\times 4$}} & \multicolumn{2}{c}{\textbf{$\times 8$}} \\
\cmidrule(lr){2-3}\cmidrule(lr){4-5}
& \textbf{PSNR ($\uparrow$)} & \textbf{SSIM ($\uparrow$)} 
& \textbf{PSNR ($\uparrow$)} & \textbf{SSIM ($\uparrow$)}  \\
\midrule
Ours        & \textbf{29.17}      & 0.650      & \textbf{27.87}      &  0.589     \\
DPS         & 25.82 & 0.548 & 22.99  & 0.507  \\
DAPS        & 28.55 & \textbf{0.690} & 27.04  & \textbf{0.630}  \\
DDS         & 28.52 & \underline{0.678} & 25.84 & \underline{0.593}  \\
FAST-DIPS   & \underline{28.79} & 0.607 & \underline{27.36} & 0.545   \\
\bottomrule
\end{tabular}
\end{table}

\subsection{Ablation studies}
\label{subsec:ablation}
Appendix~\ref{app:component_ablation} examines the effects of the
proposal rule, first-step initialization, curvature handling, and
final verification. The inertial proposal rule and conservative
first step consistently improve reconstruction quality across the
two evaluated tasks, whereas the curvature policies exhibit
task-dependent effects. Final verification does not affect the
reported reconstruction metrics in these experiments but is
retained to enforce the per-step objective nonincrease guaranteed
by Proposition~\ref{prop:pvppc_scale_displacement}.

\section{Conclusion}
\textsc{PIVOT} is a practical diffusion-prior inverse solver designed to address instability that can arise when measurement-gradient updates are inserted directly into reverse sampling. The key idea is to combine stronger heavy-ball proposal steps with objective verification, both derived from one objective for the whole trajectory{}, considered one step at a time: a candidate clean state is generated locally, evaluated against the frozen objective, and accepted only when it lowers this objective, with the denoiser prediction retained as fallback. This prevents aggressive local proposals from being reinserted when they worsen the current frozen objective.  Across FFHQ/ImageNet natural-image benchmarks, \textsc{PIVOT} provides competitive reconstruction quality{}, while it trails the best baseline in PSNR on most high dynamic range settings. Appendix~\ref{app:limit_future} discusses limitations and future directions.

\begin{ack}
This work was supported in part by the National Research Foundation of Korea (NRF) grant funded by the Korea government (MSIT) (RS-2026-25476632) and in part by Institute of Information \& communications Technology Planning \& Evaluation (IITP) under the Leading Generative AI Human Resources Development (IITP-2026-RS-2024-00360227) grant funded by the Korea government (MSIT). The authors declare no competing interests.
\end{ack}

\newpage
{
    \small
    \bibliographystyle{plain}
    \bibliography{master}
}

\newpage
\appendix
\FloatBarrier 
\clearpage 

\section{Algorithms}
\label{sec:appendix_algo}

\begin{algorithm}[H]
\caption{\textsc{PIVOT}}
\label{alg:pvppc}
\begin{algorithmic}[1]
\Require measurement \(\mathbf{y}\); differentiable operator \(A\);
clean-image map \(G\), with \(G(\mathbf{z})=\mathbf{z}\) for pixel diffusion
and \(G(\mathbf{z})=\mathrm{Dec}(\mathbf{z})\) for latent diffusion;
denoiser \(D_\sigma\); noise schedule \(\{\sigma_k\}_{k=0}^{K}\);
loss \(\ell\) with
\(\ell(\hat{\mathbf{y}},\mathbf{y})
=\|\hat{\mathbf{y}}-\mathbf{y}\|_2^2/(2\sigma_y^2)\);
number of gradient evaluations \(J\);
floor \(\sigma_{\rm floor}\);
measurement noise level \(\sigma_n\)
\State Initialize \(\mathbf{u}_0\sim\mathcal{N}(\mathbf{0},\sigma_0^2I)\)
\For{\(k=0,\ldots,K-1\)}
    \State \(\boldsymbol{\mu}_k\gets D_{\sigma_k}(\mathbf{u}_k)\)
    \State \(q_k\gets\max(\sigma_k,\sigma_{\rm floor})\),
    \(\beta_k\gets q_k^{-2}\)
    \State Define \(F(\mathbf{z})=A(G(\mathbf{z}))\)
    \State Define
    \(\Phi_k(\mathbf{z})=\ell(F(\mathbf{z}),\mathbf{y})
    +\frac{\beta_k}{2}\|\mathbf{z}-\boldsymbol{\mu}_k\|_2^2\)
    \State \(\mathbf{z}\gets\boldsymbol{\mu}_k\),
    \(\mathbf{z}_{\rm prev}\gets\mathbf{z}\)
    \State Compute data gradient
    \(\mathbf{d}\gets
    \widetilde{\nabla}_{\mathbf{z}}
    \ell(F(\mathbf{z}),\mathbf{y})\)
    \If{\(k=0\)}
        \State Initialize \(L\) by one data-gradient secant probe
        at \(\boldsymbol{\mu}_k\) \label{alg:line_probe}
    \EndIf
    \State \(w\gets L\), \(V\gets0\)
    \For{\(j=0,\ldots,J-2\)}
        \State \(\mathbf{g}\gets
        \mathbf{d}+\beta_k(\mathbf{z}-\boldsymbol{\mu}_k)\)
        \State \(M\gets w+\beta_k\);
        \(m\gets M\) if \(j=0\), otherwise \(m\gets\beta_k\)
        \State \(\alpha\gets4/(\sqrt{M}+\sqrt{m})^2\),
        \(\gamma\gets
        \big((\sqrt{M}-\sqrt{m})/(\sqrt{M}+\sqrt{m})\big)^2\)
        \State \(\mathbf{z}_{\rm new}\gets
        \mathbf{z}-\alpha\mathbf{g}
        +\gamma(\mathbf{z}-\mathbf{z}_{\rm prev})\)
        \State \(\mathbf{d}_{\rm new}\gets
        \widetilde{\nabla}_{\mathbf{z}_{\rm new}}
        \ell(F(\mathbf{z}_{\rm new}),\mathbf{y})\)
        \State \(s\gets
        \|\mathbf{d}_{\rm new}-\mathbf{d}\|_2/
        \|\mathbf{z}_{\rm new}-\mathbf{z}\|_2\)
        \Comment{guarded secant} \label{alg:line_secant}
        \State \(w\gets\max(w,s)\), \(V\gets\max(V,s)\)
        \If{\(\Phi_k(\mathbf{z}_{\rm new})>\Phi_k(\mathbf{z})\)}
            \State \(\mathbf{z}_{\rm prev}\gets\mathbf{z}_{\rm new}\),
            \(w\gets2w\)
        \Else
            \State \(\mathbf{z}_{\rm prev}\gets\mathbf{z}\)
        \EndIf
        \State \(\mathbf{z}\gets\mathbf{z}_{\rm new}\),
        \(\mathbf{d}\gets\mathbf{d}_{\rm new}\)
    \EndFor
    \If{\(\Phi_k(\mathbf{z})\;{<}
    \Phi_k(\boldsymbol{\mu}_k)\)}
        \State \(\mathbf{z}_k^+\gets\mathbf{z}\)
    \Else
        \State \(\mathbf{z}_k^+\gets\boldsymbol{\mu}_k\)
    \EndIf
    \State \(L\gets\max(V,L/2)\)
    \State Reinsert by fresh noising:
    \(\mathbf{u}_{k+1}\gets
    \mathbf{z}_k^+ + \sigma_{k+1}\boldsymbol{\epsilon}_k\),
    \(\boldsymbol{\epsilon}_k\sim\mathcal{N}(\mathbf{0},I)\)
\EndFor
\State \Return \(\hat{\mathbf{x}}=G(\mathbf{u}_K)\)
\end{algorithmic}
\end{algorithm}

Here \(\widetilde{\nabla}\ell\) denotes the data gradient returned by automatic differentiation (Appendix~\ref{app:theory_setting}),
except that output clipping uses the gated derivative of
Section~\ref{sec:method_objective_smoothing}
(Appendix~\ref{app:gate}); all evaluations of \(\Phi_k\)
use the exact forward operator.
The initial secant probe is defined in Appendix~\ref{app:theory_setting}: length \(10^{-3}\max(\|\boldsymbol{\mu}_0\|_2,10^{-6})\) along the negative data gradient, normalized with a floor of \(10^{-30}\), and \(L\) is its secant, at least \(10^{-12}\).
In line~\ref{alg:line_secant}, set \(s=0\) when the step length
is at most \(\sqrt{\varepsilon_{\rm mach}}\|\mathbf{z}\|_2\)
or the secant is nonfinite.
\FloatBarrier 
    \section{\texorpdfstring{Proofs of the theoretical properties}{Proofs of the theoretical properties}}
    \label{app:theory}
    
    This appendix gives formal assumptions and detailed proofs for the local theoretical results stated in Section~\ref{sec:method_theory}. The results of Sections~\ref{app:theory_setup}--\ref{app:polyak} are for a single fixed reverse step \(k\), with the denoiser center \(\boldsymbol{\mu}_k\), proximal weight \(\beta_k\), frozen objective \(\Phi_k\), and verified candidate set fixed. Sections~\ref{app:theory_setting}--\ref{app:theory_cond} prove the trajectory objective of Section~\ref{sec:method_overview} and the conditional results. These results do not imply global convergence of the full diffusion trajectory, because the denoiser center, noise level, reverse state, and frozen objective change from step to step.
    
    For simplicity, the proofs are written in real coordinates. Complex-valued measurements can be handled by stacking real and imaginary parts, or by replacing transposes with Hermitian adjoints where appropriate.
    
    \subsection{\texorpdfstring{Notation and the verifier}{Notation and the verifier}}
    \label{app:theory_setup}
    
    Fix a reverse step \(k\). To reduce notation, write
    \[
        \boldsymbol{\mu}=\boldsymbol{\mu}_k,
        \qquad
        \beta=\beta_k,
        \qquad
        L(\mathbf z)=\ell(F(\mathbf z),\mathbf y).
    \]
    The exact frozen objective is
    \begin{equation}
        \Phi(\mathbf z)
        =
        L(\mathbf z)
        +
        \frac{\beta}{2}\|\mathbf z-\boldsymbol{\mu}\|_2^2.
        \label{eq:app_phi}
    \end{equation}

    The verified candidate set is

    \begin{equation}
        \mathcal S=\{\boldsymbol{\mu},\mathbf z^{(J-1)}\}.
        \label{eq:app_candidate_set_new}
    \end{equation}
    The exact best-candidate verifier returns
    \begin{equation}
        \mathbf z^+
        \in
        \arg\min_{\mathbf z\in\text{$\mathcal S$}}\Phi(\mathbf z).
        \label{eq:app_exact_verifier}
    \end{equation}
    Since \(\boldsymbol{\mu}\in\text{$\mathcal S$}\), exact verification implies
    \[
        \Phi(\mathbf z^+)\leq \Phi(\boldsymbol{\mu}).
    \]

    The method commits \(\mathbf z^{(J-1)}\) only if \(\Phi(\mathbf z^{(J-1)})<\Phi(\boldsymbol{\mu})\), so it is exact and takes the anchor in a tie.

    \subsection{Proof of Proposition~\ref{prop:pvppc_scale_displacement}}
    \label{app:proof_scale_displacement}
    
    We prove the nonincrease claim and the displacement and reinsertion bounds.

    \subsubsection{Nonincrease}
    
    The candidate set \(\mathcal S=\{\boldsymbol{\mu},\mathbf z^{(J-1)}\}\) has at most two elements, and exact verification returns a minimizer of \(\Phi\) over it, so \(\Phi(\mathbf z^+)=\min\{\Phi(\boldsymbol{\mu}),\Phi(\mathbf z^{(J-1)})\}\leq\Phi(\boldsymbol{\mu})=L(\boldsymbol{\mu})\), whatever produced \(\mathbf z^{(J-1)}\).
    
    \subsubsection{Displacement control}
    
    Let \(\boldsymbol{\Delta}^+=\mathbf z^+-\boldsymbol{\mu}\). Because \(\boldsymbol{\mu}\in\mathcal S\), exact verification gives
    \[
        \Phi(\mathbf z^+)\leq\Phi(\boldsymbol{\mu}).
    \]

    Expanding the definition of \(\Phi\),
    \[
        \Phi(\mathbf z^+)
        =
        L(\mathbf z^+)
        +
        \frac{\beta}{2}\|\mathbf z^+-\boldsymbol{\mu}\|_2^2
        =
        L(\mathbf z^+)
        +
        \frac{\beta}{2}\|\boldsymbol{\Delta}^+\|_2^2.
    \]
    At the baseline,
    \[
        \Phi(\boldsymbol{\mu})
        =
        L(\boldsymbol{\mu})
        +
        \frac{\beta}{2}\|\boldsymbol{\mu}-\boldsymbol{\mu}\|_2^2
        =
        L(\boldsymbol{\mu}).
    \]
    Therefore,
    \[
        L(\mathbf z^+)
        +
        \frac{\beta}{2}\|\boldsymbol{\Delta}^+\|_2^2
        \leq
        L(\boldsymbol{\mu}).
    \]
    Rearranging,
    \[
        \frac{\beta}{2}\|\boldsymbol{\Delta}^+\|_2^2
        \leq
        L(\boldsymbol{\mu})-L(\mathbf z^+).
    \]
    
    \subsubsection{\texorpdfstring{Fresh-noise reinsertion: KL}{Fresh-noise reinsertion: KL}}
    
    For fresh-noise reinsertion with \(\sigma_{k+1}>0\), define
    \[
        Q^+
        =
        \mathcal N(\mathbf z^+,\sigma_{k+1}^2I),
        \qquad
        Q^0
        =
        \mathcal N(\boldsymbol{\mu},\sigma_{k+1}^2I).
    \]
    These Gaussian distributions have the same covariance and means \(\mathbf z^+\) and \(\boldsymbol{\mu}\).

    Their densities at \(\mathbf x\) are proportional, with the same constant, to \(\exp(-\|\mathbf x-\mathbf z^+\|_2^2/(2\sigma_{k+1}^2))\) and \(\exp(-\|\mathbf x-\boldsymbol{\mu}\|_2^2/(2\sigma_{k+1}^2))\), so
    \[
        \log\frac{dQ^+}{dQ^0}(\mathbf x)
        =
        \frac{\|\mathbf x-\boldsymbol{\mu}\|_2^2-\|\mathbf x-\mathbf z^+\|_2^2}{2\sigma_{k+1}^2}.
    \]
    Under \(Q^+\), write \(\mathbf x=\mathbf z^++\sigma_{k+1}\boldsymbol{\xi}\) with \(\boldsymbol{\xi}\sim\mathcal N(\mathbf 0,I)\). Then \(\mathbf x-\boldsymbol{\mu}=\boldsymbol{\Delta}^++\sigma_{k+1}\boldsymbol{\xi}\) and \(\mathbf x-\mathbf z^+=\sigma_{k+1}\boldsymbol{\xi}\), so
    \[
    \begin{aligned}
        \|\mathbf x-\boldsymbol{\mu}\|_2^2-\|\mathbf x-\mathbf z^+\|_2^2
        &=\|\boldsymbol{\Delta}^+\|_2^2+2\sigma_{k+1}(\boldsymbol{\Delta}^+)^\top\boldsymbol{\xi}+\sigma_{k+1}^2\|\boldsymbol{\xi}\|_2^2-\sigma_{k+1}^2\|\boldsymbol{\xi}\|_2^2\\
        &=\|\boldsymbol{\Delta}^+\|_2^2+2\sigma_{k+1}(\boldsymbol{\Delta}^+)^\top\boldsymbol{\xi}.
    \end{aligned}
    \]
    Taking the expectation under \(Q^+\), which defines \(\mathrm{KL}(Q^+\|Q^0)=\mathbb E_{Q^+}[\log(dQ^+/dQ^0)]\), and using \(\mathbb E[\boldsymbol{\xi}]=\mathbf 0\),
    \[
        \mathrm{KL}(Q^+\|Q^0)
        =
        \frac{\|\boldsymbol{\Delta}^+\|_2^2}{2\sigma_{k+1}^2}.
    \]
    Using the displacement-control inequality,
    \[
        \|\boldsymbol{\Delta}^+\|_2^2
        \leq
        \frac{2}{\beta}
        [L(\boldsymbol{\mu})-L(\mathbf z^+)].
    \]
    Therefore,
    \[
        \mathrm{KL}(Q^+\|Q^0)
        \leq
        \frac{
        L(\boldsymbol{\mu})-L(\mathbf z^+)
        }{
        \beta\sigma_{k+1}^2
        }.
    \]
    This completes the proof of Proposition~\ref{prop:pvppc_scale_displacement}.

    \subsection{Step size and momentum of the proposal}
    \label{app:polyak}
    Section~\ref{sec:method_overview} sets the step size and momentum of the heavy-ball proposal from a model in which the curvature of
    \(\Phi_k\) lies in \([m,M]\), with \(m=\beta_k\) (the proximal weight) and \(M=\beta_k+w\) (\(w\) the curvature model of the data term). We derive both choices.

    \paragraph{First step.}
    Let \(\Phi\) be differentiable, \(\mathbf g=\nabla\Phi(\mathbf z)\), \(\alpha>0\), and \(\|\nabla\Phi(\mathbf a)-\nabla\Phi(\mathbf b)\|_2\le M\|\mathbf a-\mathbf b\|_2\) for
    \(\mathbf a,\mathbf b\) on the segment from \(\mathbf z\) to \(\mathbf z-\alpha\mathbf g\). Then, by the fundamental theorem of calculus along the segment (its integrand is continuous because \(\nabla\Phi\) is Lipschitz there),
    \[
    \begin{aligned}
        \Phi(\mathbf z-\alpha\mathbf g)-\Phi(\mathbf z)
        &=\int_0^1\nabla\Phi(\mathbf z-t\alpha\mathbf g)^\top(-\alpha\mathbf g)\,dt\\
        &=-\alpha\|\mathbf g\|_2^2-\alpha\int_0^1\big(\nabla\Phi(\mathbf z-t\alpha\mathbf g)-\nabla\Phi(\mathbf z)\big)^\top\mathbf g\,dt\\
        &\le-\alpha\|\mathbf g\|_2^2+\alpha\int_0^1Mt\alpha\|\mathbf g\|_2^2\,dt
        =-\alpha\Big(1-\frac{\alpha M}{2}\Big)\|\mathbf g\|_2^2,
    \end{aligned}
    \]
    where the inequality is the Cauchy--Schwarz inequality with \(\|\nabla\Phi(\mathbf z-t\alpha\mathbf g)-\nabla\Phi(\mathbf z)\|_2\le Mt\alpha\|\mathbf g\|_2\).
    The guaranteed decrease \(\alpha(1-\alpha M/2)\|\mathbf g\|_2^2\) is a concave quadratic in \(\alpha\) with derivative \((1-\alpha M)\|\mathbf g\|_2^2\), so it is
    largest at \(\alpha=1/M\), where it equals \(\|\mathbf g\|_2^2/(2M)\).

    \paragraph{Heavy-ball steps.}
    Let \(\Phi(\mathbf z)=\frac12(\mathbf z-\mathbf z^\star)^\top\mathbf C(\mathbf z-\mathbf z^\star)\) with \(\mathbf C\) symmetric and its eigenvalues in \([m,M]\), \(0<m<M\), and
    \(\mathbf z^{(j+1)}=\mathbf z^{(j)}-\alpha\nabla\Phi(\mathbf z^{(j)})+\gamma(\mathbf z^{(j)}-\mathbf z^{(j-1)})\) with \(\alpha>0\) and \(0\le\gamma<1\). The error
    \(\mathbf h^{(j)}=\mathbf z^{(j)}-\mathbf z^\star\) satisfies \(\mathbf h^{(j+1)}=((1+\gamma)I-\alpha\mathbf C)\mathbf h^{(j)}-\gamma\mathbf h^{(j-1)}\), so its coordinate along an
    eigenvector of \(\mathbf C\) with eigenvalue \(a\) obeys \(h_{j+1}=s(a)h_j-\gamma h_{j-1}\), \(s(a)=1+\gamma-\alpha a\). The worst-case asymptotic rate of this coordinate over its initial values is the spectral factor, the
    largest modulus \(\varrho(a)\) of the roots \(r_1,r_2\) of \(p_a(r)=r^2-s(a)r+\gamma\), for which \(r_1r_2=\gamma\) and \(r_1+r_2=s(a)\).

    \emph{(i) Polyak's choice.} Let \(q^\star=(\sqrt M-\sqrt m)/(\sqrt M+\sqrt m)\), \(\gamma=(q^\star)^2\) and \(\alpha=4/(\sqrt M+\sqrt m)^2\). Since
    \(1-q^\star=2\sqrt m/(\sqrt M+\sqrt m)\) and \(1+q^\star=2\sqrt M/(\sqrt M+\sqrt m)\), \(\alpha m=(1-q^\star)^2\) and \(\alpha M=(1+q^\star)^2\). For \(a\in[m,M]\) this gives
    \((1-\sqrt\gamma)^2\le\alpha a\le(1+\sqrt\gamma)^2\), that is \(-2\sqrt\gamma\le s(a)\le2\sqrt\gamma\), so the discriminant \(s(a)^2-4\gamma\) is not positive: the roots are
    complex conjugate or double, and \(|r_1|=|r_2|=\sqrt{r_1r_2}=\sqrt\gamma\). Hence \(\varrho(a)=q^\star\) for every \(a\in[m,M]\).

    \emph{(ii) No choice does better at both ends.} Let \(\alpha>0\), \(0\le\gamma<1\) and \(q<1\) with \(\varrho(m)\le q\) and \(\varrho(M)\le q\). Then \(q>0\): \(\varrho(m)=\varrho(M)=0\)
    would need \(s(m)=s(M)=0\), that is \(\alpha m=\alpha M\). For \(a\in\{m,M\}\), if the roots are complex, \(\gamma=|r_1|^2\le q^2\) and
    \(|s(a)|<2\sqrt\gamma\le q+\gamma/q\), the last step being \((\sqrt q-\sqrt{\gamma/q})^2\ge0\); if they are real, both lie in \([-q,q]\), so \(\gamma=r_1r_2\le q^2\),
    \(p_a(q)=(q-r_1)(q-r_2)\ge0\) and \(p_a(-q)=(q+r_1)(q+r_2)\ge0\), and dividing \(q^2\mp s(a)q+\gamma\ge0\) by \(q\) gives \(|s(a)|\le q+\gamma/q\). In both cases
    \(\gamma\le q^2\) and \(|s(a)|\le q+\gamma/q\). At \(a=m\), \(s(m)\le q+\gamma/q\) gives \(\alpha m\ge1+\gamma-q-\gamma/q=(1-q)(1-\gamma/q)\); at \(a=M\),
    \(s(M)\ge-(q+\gamma/q)\) gives \(\alpha M\le1+\gamma+q+\gamma/q=(1+q)(1+\gamma/q)\). Since \(\gamma/q\le q<1\), \((1-q)(1-\gamma/q)>0\), and dividing,
    \[
        \frac Mm\le\frac{(1+q)(1+\gamma/q)}{(1-q)(1-\gamma/q)}\le\Big(\frac{1+q}{1-q}\Big)^2,
    \]
    because \(x\mapsto(1+x)/(1-x)\) increases on \([0,1)\) and \(\gamma/q\le q\). Taking square roots, \(\sqrt{M/m}\le(1+q)/(1-q)\), that is
    \(q\ge(\sqrt{M/m}-1)/(\sqrt{M/m}+1)=q^\star\).

    Hence Polyak's step size and momentum give every curvature in \([m,M]\) the worst-case asymptotic rate \(q^\star\), a spectral factor rather than a one-step contraction, and no step size and momentum give a smaller rate at both
    \(m\) and \(M\). If \(m=M\) (that is \(w=0\)), \(\mathbf C=MI\), the formulas give \(\gamma=0\) and \(\alpha=1/M\), and one step reaches \(\mathbf z^\star\). \textsc{PIVOT} uses them with \(m=\beta_k\) and \(M=\beta_k+w\); its first step, which has no momentum term, uses \(\alpha=1/M\).

    \subsection{Setting for the whole-run results}
    \label{app:theory_setting}

    \paragraph{Steps and objective.}
    The sampler runs $K$ reverse steps with EDM noise levels $\sigma_0>\dots>\sigma_{K-1}>\sigma_K=0$,
    $\sigma_i=\big(\sigma_{\max}^{1/7}+\frac iK(\sigma_{\min}^{1/7}-\sigma_{\max}^{1/7})\big)^7$ for $0\le i\le K-1$
    ($\sigma_{\max}=80$ in pixel and $10$ in latent space, $\sigma_{\min}=0.02$). For every $K$,
    $\sigma_{K-1}=\big(\sigma_{\min}^{1/7}+(\sigma_{\max}^{1/7}-\sigma_{\min}^{1/7})/K\big)^7>\sigma_{\min}$, above the floor
    $\sigma_{\rm floor}=10^{-3}$, so $\beta_k=\sigma_k^{-2}$ exactly; write $\rho_k=\sigma_{k+1}/\sigma_k$ for $k\le K-2$. The
    state is $\pvu_k$, with $\pvu_0=\sigma_0\pveps_{-1}$, and the anchor is $\pvmu_k=D_{\sigma_k}(\pvu_k)$ for a deterministic denoiser
    $D_\sigma$. With the Gaussian loss,
    \begin{equation}\label{eq:th_obj}
      L(\pvz)=\frac{\|F(\pvz)-\pvy\|^2}{2\sigma_y^2},\qquad
      \Phi_k(\pvz)=L(\pvz)+\frac{\beta_k}{2}\|\pvz-\pvmu_k\|^2,\qquad \sigma_y>0 .
    \end{equation}
    The implementation replaces a non-finite value of $L$ by $+\infty$ before every comparison (the restart test and the verifier), so $\Phi_k(\pvmu_k)=L(\pvmu_k)\in[0,+\infty]$; $\Phi_k$ of a candidate is not a number only if the candidate itself has non-finite entries, and such a candidate is never committed, because the strict comparison with it is false. If $L(\pvmu_k)=+\infty$ we set $\Delta L_k:=+\infty$ below; every bound
    that involves $\Delta L_k$ is then trivial.

    \paragraph{The field the steps follow.}
    The steps use the field $\pvg(\pvz)=\pvg_L(\pvz)+\beta_k(\pvz-\pvmu_k)$, where $\pvg_L(\pvz)$ is the gradient of $L$ at $\pvz$ returned by
    automatic differentiation, with the gated derivative of Section~\ref{sec:method_objective_smoothing} for a clip at the output of the forward model (non-finite entries, if any, set to zero), and the verifier evaluates $\Phi_k$ itself. At a non-differentiable primitive, automatic differentiation applies that primitive's prescribed backward rule, so $\pvg_L$ need not be a classical or a one-sided derivative of $L$ there; $\pvg_L=\nabla L$ where the evaluated computational graph satisfies the classical chain rule and the gate has not modified it, and $\pvg_L=\nabla\tilde L$ (Section~\ref{app:gate}) where the graph satisfies the chain rule and the gate releases coordinates. The results of Sections~\ref{app:theory_setting}--\ref{app:theory_cond} use one property of $\pvg_L$ only: it is a
    Borel function of $\pvz$ (for fixed $\pvy$), because automatic differentiation evaluates finitely many primitive backward maps,
    each continuous on each of finitely many Borel pieces selected by comparisons (the same reason makes $D_\sigma$ and $L$ Borel).
    No result uses that $\pvg_L$ is a gradient, is continuous, or is a descent direction of $\Phi_k$.

    \paragraph{One step.}
    (1) \emph{Proposal.} From $\pvz_k^{(0)}=\pvmu_k$, the $J-1$ heavy-ball steps \eqref{eq:hb_update} on the field $\pvg$ use per-image scalars
    $\alpha$, $\gamma$ set by Polyak's formula from the curvature model $w$, which starts at the curvature $w_k$ carried into the step,
    is raised to every guarded secant observed in the step and doubled at every function-value restart; the curvature carried to
    the next step is $w_{k+1}=\max(V_k,w_k/2)$, with $V_k$ the largest guarded secant of the step, and $w_0$ is the initial secant along one
    probe step from $\pvmu_0$: the probe has length $10^{-3}\max(\|\pvmu_0\|,10^{-6})$ along $-\pvg_L(\pvmu_0)/\max(\|\pvg_L(\pvmu_0)\|,10^{-30})$, and $w_0=\max(s_0,10^{-12})$, where $s_0$ is its secant with the displacement norm floored at $10^{-30}$, so $w_0$ is defined for every anchor (a zero gradient gives $w_0=10^{-12}$) (Algorithm~\ref{alg:pvppc}). The end point $\pvz_k^{(J-1)}$ is the \emph{candidate}. No result below uses
    the form of the steps, the values of $\alpha$, $\gamma$, the restart rule or the carry: Lemma~\ref{lem:traj_kl} (and through it Corollary~\ref{cor:trajectory_objective}) uses only
    that $w_{k+1}$ and $\pvz_k^{(J-1)}$ are functions of $(\pvu_k,w_k)$, and every other result holds for every value of $\pvz_k^{(J-1)}$.
    (2) \emph{Verification.} The commit \eqref{eq:best_candidate_verification} is the first strict argmin of $\Phi_k$ over
    $\{\pvmu_k,\pvz_k^{(J-1)}\}$, the anchor first; it is the exact verifier of Section~\ref{app:theory_setup}. We write
    \[
      \pvDelta_k=\pvz^+_k-\pvmu_k,\quad T_k=\frac{\beta_k}{2}\|\pvDelta_k\|^2,\quad
      \Delta L_k=L(\pvmu_k)-L(\pvz^+_k),\quad
      \pvKL_k=\frac{\|\pvDelta_k\|^2}{2\sigma_{k+1}^2}\ \ (k\le K-2).
    \]
    By Proposition~\ref{prop:pvppc_scale_displacement}, at every step
    \begin{equation}\label{eq:th_cert}
      \Phi_k(\pvz^+_k)\le L(\pvmu_k),\qquad T_k\le\Delta L_k,
    \end{equation}
    and at a step whose commit is the anchor, $\pvDelta_k=\mathbf 0$, $T_k=0$ and, when $L(\pvmu_k)<\infty$, $\Delta L_k=0$.

    \paragraph{Transition.}
    For $k\le K-2$ the reinsertion noise is $\pveps_k\sim\pvN(0,I)$, and
    \begin{equation}\label{eq:th_transition}
      \pvu_{k+1}=\pvz^+_k+\sigma_{k+1}\pveps_k\quad(0\le k\le K-2),\qquad \pvu_K=\pvz^+_{K-1}\ (\sigma_K=0),\qquad \hat\pvx=G(\pvu_K)=G(\pvz^+_{K-1}).
    \end{equation}

    \paragraph{Probability model.}
    $\pveps_{-1},\pveps_0,\dots,\pveps_{K-2}$ are independent standard Gaussian vectors; $\mathbb E$ denotes expectation over them, with
    $\pvy$ fixed. \emph{Functional dependence.} At $k=0$ the carried curvature $w_0$ is a function of $\pvmu_0=D_{\sigma_0}(\pvu_0)$. The
    computation of step $k$ (the proposal with its secants, restarts and curvature model, the verifier and the carry) is a
    deterministic function of $(\pvu_k,w_k)$: $\pvmu_k=D_{\sigma_k}(\pvu_k)$, and every later quantity is obtained from $\pvmu_k$, $w_k$ and
    the fixed $\pvy$ by a fixed finite number of evaluations of $D_\sigma$, $L$ and $\pvg_L$, arithmetic operations, maxima and
    comparisons. Hence, by induction on $k$, $w_0$ is a function of $\pvu_0$, $w_k$ is a function of $\pvu_{0:k-1}$ for $k\ge1$, and
    $\pvz^+_k=\zeta_k(\pvu_{0:k})$ for a map $\zeta_k:\mathbb R^{d(k+1)}\to\mathbb R^d$. \emph{Measurability.} $\zeta_k$ is Borel: $D_\sigma$,
    $L$ and $\pvg_L$ are Borel, arithmetic operations and maxima are continuous, a selection by comparison
    $\mathbf x\mapsto\mathbf 1[\varphi(\mathbf x)<\psi(\mathbf x)]\,\mathbf a(\mathbf x)+\mathbf 1[\varphi(\mathbf x)\ge\psi(\mathbf x)]\,\mathbf b(\mathbf x)$ with Borel
    $\varphi,\psi,\mathbf a,\mathbf b$ is Borel, and a finite composition of Borel maps is Borel. \emph{Independence.} By
    \eqref{eq:th_transition} and induction, $\pvu_{0:k}$ is a function of $(\pveps_{-1},\dots,\pveps_{k-1})$, and $\pveps_k$ is independent of
    these; hence $\pveps_k$ is independent of $\pvu_{0:k}$. All statements are per image and in exact arithmetic; the denoiser is
    idealised as one deterministic map with finite values.

    \paragraph{Standard facts used in the proofs.}
    \begin{enumerate}
    \item[(F1)] (Gaussian KL.) For $\mathbf a,\mathbf b\in\mathbb R^d$ and $s>0$,
      $\pvKL(\pvN(\mathbf a,s^2I)\|\pvN(\mathbf b,s^2I))=\|\mathbf a-\mathbf b\|^2/(2s^2)$. This is derived in Appendix~\ref{app:proof_scale_displacement} with \(\mathbf a=\mathbf z^+\), \(\mathbf b=\boldsymbol{\mu}\) and \(s=\sigma_{k+1}\); the derivation uses only that both laws are Gaussian with covariance \(s^2I\).
    \item[(F2)] (Chain rule.) Let $P=P_X\otimes P_{Y\mid X}$ and $Q=Q_X\otimes Q_{Y\mid X}$ be laws of a pair $(X,Y)$ of random vectors,
      built from laws $P_X,Q_X$ of $X$ and Markov kernels $P_{Y\mid X},Q_{Y\mid X}$ such that, for every \(\mathbf x\), \(P_{Y\mid X=\mathbf x}\) has a density \(g(\mathbf x,\cdot)\) with respect to \(Q_{Y\mid X=\mathbf x}\), with \(g\) jointly Borel. Then, in $[0,\infty]$,
      $\pvKL(P\|Q)=\pvKL(P_X\|Q_X)+\int\pvKL(P_{Y\mid X=\mathbf x}\|Q_{Y\mid X=\mathbf x})\,P_X(d\mathbf x)$. Derivation: if $P_X\not\ll Q_X$, a Borel set $A$ with $Q_X(A)=0<P_X(A)$ gives $Q(X\in A)=Q_X(A)=0<P_X(A)=P(X\in A)$, so $P\not\ll Q$ and both sides are $+\infty$.
      Otherwise let $f=dP_X/dQ_X$. By Tonelli's theorem, $P(E)=\int\!\int\mathbf 1_E(\mathbf x,\mathbf y)\,g(\mathbf x,\mathbf y)\,Q_{Y\mid X=\mathbf x}(d\mathbf y)\,f(\mathbf x)\,Q_X(d\mathbf x)$ for every Borel $E$, so
      $dP/dQ(\mathbf x,\mathbf y)=f(\mathbf x)g(\mathbf x,\mathbf y)$ and $\pvKL(P\|Q)=\int[\log f(\mathbf x)+\log g(\mathbf x,\mathbf y)]\,P(d\mathbf x,d\mathbf y)$. Since $t\log(1/t)\le1/e$ for $t\in(0,1]$ (its derivative $\log(1/t)-1$ is positive for $t<1/e$ and negative for $t>1/e$, and its value at $t=1/e$ is $1/e$), the negative parts are
      integrable: $\int(\log f)^-dP_X=\int_{\{f<1\}}f\log(1/f)\,dQ_X\le1/e$, and likewise $\int(\log g(\mathbf x,\cdot))^-dP_{Y\mid X=\mathbf x}\le1/e$ for every $\mathbf x$. Hence the integral is the sum of
      $\int\log f\,dP_X=\pvKL(P_X\|Q_X)$ and, by Tonelli's theorem applied to the positive and the negative part,
      $\int\big[\int\log g(\mathbf x,\mathbf y)\,P_{Y\mid X=\mathbf x}(d\mathbf y)\big]P_X(d\mathbf x)=\int\pvKL(P_{Y\mid X=\mathbf x}\|Q_{Y\mid X=\mathbf x})\,P_X(d\mathbf x)$.
    
    \end{enumerate}

    \subsection{\texorpdfstring{One objective for the trajectory}{One objective for the trajectory}}
    \label{app:theory_budget}

    Let $P$ be the law of $(\pvu_0,\dots,\pvu_{K-1})$ under the probability model of Section~\ref{app:theory_setting}, and $Q$ the law of
    the \emph{uncorrected chain} $\pvv_0\overset{d}{=}\pvu_0$, $\pvv_{k+1}=D_{\sigma_k}(\pvv_k)+\sigma_{k+1}\pveps_k$ with $\pveps_k$ independent
    standard Gaussians (a reference that is never run).

    \begin{lemma}[Trajectory KL identity]\label{lem:traj_kl}
    In $[0,\infty]$, $\pvKL(P\|Q)=\sum_{k\le K-2}\mathbb E\big[\|\pvDelta_k\|^2/(2\sigma_{k+1}^2)\big]$, the expectation being over $\pveps_{-1},\dots,\pveps_{K-2}$ with $\pvy$ fixed.
    \end{lemma}
    \vspace{0pt}

    \begin{proof}

     \emph{Step 1 (one transition of the run).} Fix $k\le K-2$. By \eqref{eq:th_transition} and the probability model,
    $\pvu_{k+1}=\zeta_k(\pvu_{0:k})+\sigma_{k+1}\pveps_k$, where $\zeta_k$ is Borel, $\pvz^+_k=\zeta_k(\pvu_{0:k})$, and
    $\pveps_k\sim\pvN(0,I)$ is independent of $\pvu_{0:k}$.

    \emph{Step 2 (conditional law under $P$).} Write $\varphi_d$ for the standard Gaussian law on $\mathbb R^d$ and note that
    $\pvN(\mathbf a,s^2I)(E)=\int\mathbf 1_E(\mathbf a+s\pveps)\,\varphi_d(d\pveps)$ for Borel $E$. For Borel $E\subset\mathbb R^d$ and
    $E'\subset\mathbb R^{d(k+1)}$, by Step 1, the independence of $\pveps_k$ and $\pvu_{0:k}$ and Fubini's theorem on their product law,
    \[
      \begin{aligned}
      P(\pvu_{0:k}\in E',\ \pvu_{k+1}\in E)&=\mathbb E\big[\mathbf 1_{E'}(\pvu_{0:k})\,\mathbf 1_E(\zeta_k(\pvu_{0:k})+\sigma_{k+1}\pveps_k)\big]\\
      &=\mathbb E\big[\mathbf 1_{E'}(\pvu_{0:k})\,\pvN(\zeta_k(\pvu_{0:k}),\sigma_{k+1}^2I)(E)\big].
      \end{aligned}
    \]
    Hence the Markov kernel $\pvu_{0:k}\mapsto\pvN(\zeta_k(\pvu_{0:k}),\sigma_{k+1}^2I)$ (a kernel because $\zeta_k$ is Borel) is a
    regular conditional law of $\pvu_{k+1}$ given $\pvu_{0:k}$ under $P$:
    $P(\pvu_{k+1}\in\cdot\mid\pvu_{0:k})=\pvN(\pvz^+_k,\sigma_{k+1}^2I)$.

    \emph{Step 3 (conditional law under $Q$).} Under $Q$, $\pvv_{0:k}$ is a function of $(\pveps_{-1},\dots,\pveps_{k-1})$ and
    $\pvv_{k+1}=D_{\sigma_k}(\pvv_k)+\sigma_{k+1}\pveps_k$ with $\pveps_k$ independent of $\pvv_{0:k}$ ($D_\sigma$ is Borel), so the same
    computation gives $Q(\pvv_{k+1}\in\cdot\mid\pvv_{0:k})=\pvN(D_{\sigma_k}(\pvv_k),\sigma_{k+1}^2I)$; evaluated at $\pvv_{0:k}=\pvu_{0:k}$ this is
    $\pvN(D_{\sigma_k}(\pvu_k),\sigma_{k+1}^2I)=\pvN(\pvmu_k,\sigma_{k+1}^2I)$.

    \emph{Step 4 (chain rule).} Let $P^{(j)}$ and $Q^{(j)}$ be the laws of $\pvu_{0:j}$ under $P$ and of $\pvv_{0:j}$ under $Q$, so that
    $P=P^{(K-1)}$ and $Q=Q^{(K-1)}$. By Steps 2--3, $P^{(k+1)}=P^{(k)}\otimes\pvN(\zeta_k(\cdot),\sigma_{k+1}^2I)$ and
    $Q^{(k+1)}=Q^{(k)}\otimes\pvN(D_{\sigma_k}((\cdot)_k),\sigma_{k+1}^2I)$, where $(\cdot)_k$ is the last block of the path. (F2) with
    $X=\pvu_{0:k}$ and $Y=\pvu_{k+1}$ (its hypothesis holds: the density of \(\pvN(\zeta_k(\pvu_{0:k}),\sigma_{k+1}^2I)\) with respect to \(\pvN(D_{\sigma_k}(\pvu_k),\sigma_{k+1}^2I)\) at \(\mathbf y\) is \(\exp\big((\|\mathbf y-D_{\sigma_k}(\pvu_k)\|^2-\|\mathbf y-\zeta_k(\pvu_{0:k})\|^2)/(2\sigma_{k+1}^2)\big)\), a Borel function of \((\pvu_{0:k},\mathbf y)\)) gives
    \[
      \pvKL(P^{(k+1)}\|Q^{(k+1)})=\pvKL(P^{(k)}\|Q^{(k)})+\mathbb E\,\pvKL\Big(\pvN(\pvz^+_k,\sigma_{k+1}^2I)\,\Big\|\,\pvN(\pvmu_k,\sigma_{k+1}^2I)\Big),
    \]
    the expectation being over $\pvu_{0:k}\sim P^{(k)}$. Adding these identities for $k=0,\dots,K-2$,
    \[
      \pvKL(P\|Q)=\pvKL(P^{(0)}\|Q^{(0)})+\sum_{k=0}^{K-2}\mathbb E\,\pvKL\Big(\pvN(\pvz^+_k,\sigma_{k+1}^2I)\,\Big\|\,\pvN(\pvmu_k,\sigma_{k+1}^2I)\Big),
    \]
    and the first term is $0$ because $\pvv_0\overset{d}{=}\pvu_0$. (The versions of the kernels do not matter: $P^{(0)}=Q^{(0)}$, and
    two Gaussian laws with the same covariance are mutually absolutely continuous, so $P^{(k)}\ll Q^{(k)}$ for every $k$ by induction.)

    \emph{Step 5 (one term).} For every value of $\pvu_{0:k}$, (F1) with $\mathbf a=\pvz^+_k$, $\mathbf b=\pvmu_k$ and $s=\sigma_{k+1}$ gives
    $\pvKL\big(\pvN(\pvz^+_k,\sigma_{k+1}^2I)\|\pvN(\pvmu_k,\sigma_{k+1}^2I)\big)=\|\pvDelta_k\|^2/(2\sigma_{k+1}^2)=\pvKL_k$.

    \emph{Step 6 (sum).} Inserting Step 5 in Step 4 gives $\pvKL(P\|Q)=\sum_{k=0}^{K-2}\mathbb E[\pvKL_k]$, the claim, because $\pvKL_k=\|\pvDelta_k\|^2/(2\sigma_{k+1}^2)$.

    \end{proof}

    \begin{corollary}[One objective for the trajectory]\label{cor:trajectory_objective}
    With $\lambda_k=\rho_k^{-2}=\sigma_k^2/\sigma_{k+1}^2$,
    \[
      \mathbb E\Big[\sum_{k\le K-2}\lambda_kL(\pvz^+_k)\Big]+\pvKL(P\|Q)=\sum_{k\le K-2}\lambda_k\,\mathbb E\big[\Phi_k(\pvz^+_k)\big]\qquad\text{in }[0,\infty].
    \]
    \end{corollary}

    \begin{proof}
    Lemma~\ref{lem:traj_kl} gives $\pvKL(P\|Q)=\sum_{k\le K-2}\mathbb E[\pvKL_k]$, and $\pvKL_k=\|\pvDelta_k\|^2/(2\sigma_{k+1}^2)=\lambda_k\beta_k\|\pvDelta_k\|^2/2$
    because $\beta_k=\sigma_k^{-2}$. Adding $\sum_{k\le K-2}\lambda_k\mathbb E[L(\pvz^+_k)]$ (all terms in $[0,\infty]$, linearity of the expectation) and using
    $\Phi_k(\pvz^+_k)=L(\pvz^+_k)+\frac{\beta_k}{2}\|\pvDelta_k\|^2$ from \eqref{eq:th_obj} gives the claim.
    \end{proof}

    \subsection{Conditional results}
    \label{app:theory_cond}

    Proposition~\ref{prop:cond_quality} and Theorem~\ref{thm:cond_recursion} relate the error of the committed point to that of the anchor and
    follow it along the run; Proposition~\ref{prop:cond_coupling} bounds the distance to the uncorrected chain.  The
    committed point obeys \eqref{eq:th_cert} at every step. Every hypothesis below is an assumption about the operator, the denoiser
    or the ground truth; none is enforced by the method or measured. Throughout, $L(\pvmu_k)<\infty$, so that $r_k$ and $\theta_k$ below are finite.

    \begin{assumption}[Bounded noise]\label{asm:th_noise}
    $\pvy=A(\pvx^\star)+\pvn$ with $\|\pvn\|\le\delta$.
    \end{assumption}

    \begin{assumption}[Restricted measurement stability at step $k$]\label{asm:th_stability}
    With $\pvx_k=G(\pvmu_k)$ and $\tilde\pvx_k=G(\pvz^+_k)$:
    $m_k\|\tilde\pvx_k-\pvx^\star\|\le\|A(\tilde\pvx_k)-A(\pvx^\star)\|$ and $\|A(\pvx_k)-A(\pvx^\star)\|\le M_k\|\pvx_k-\pvx^\star\|$,
    with $0<m_k$ and $0<M_k<\infty$.
    \end{assumption}
    This is a two-point condition, weaker than global injectivity; its content is in the values of $m_k$ and $M_k$. For linear $A$,
    $m_k$ is small when the error of $\tilde\pvx_k$ lies mostly in a null space of the operator (super resolution, inpainting,
    frequencies a blur removes or strongly attenuates), and no $m_k>0$ exists when it lies entirely in one. For phase retrieval
    the same happens near a twin or a global sign flip.

    \begin{assumption}[Transition stability]\label{asm:th_transition}
    $\|\pvx_{k+1}-\pvx^\star\|\le c_{k+1}\|\tilde\pvx_k-\pvx^\star\|+b_{k+1}$ with $c_{k+1},b_{k+1}\ge0$, where
    $\pvx_{k+1}=G\big(D_{\sigma_{k+1}}(\pvz^+_k+\sigma_{k+1}\pveps_k)\big)$ is the image of the next anchor.
    \end{assumption}

    \begin{assumption}[Denoiser Lipschitz on the coupled pair]\label{asm:th_lipschitz}
    $\|D_{\sigma_k}(\pvu_k)-D_{\sigma_k}(\pvv_k)\|\le\Lambda_k\|\pvu_k-\pvv_k\|$ with $0\le\Lambda_k<\infty$, where $\pvv_k$ is the uncorrected chain driven by
    the run's noise: $\pvv_0=\pvu_0$ and $\pvv_{k+1}=D_{\sigma_k}(\pvv_k)+\sigma_{k+1}\pveps_k$ for $k\le K-2$, and $\pvv_K=D_{\sigma_{K-1}}(\pvv_{K-1})$ since $\sigma_K=0$.
    \end{assumption}

    Write $r(\pvz)=\|F(\pvz)-\pvy\|$, $r_k=r(\pvmu_k)$, $e_k=\|\pvx_k-\pvx^\star\|$ and
    \begin{equation}\label{eq:th_theta}
      \theta_k=\Big(1-\frac{\beta_k\sigma_y^2\|\pvDelta_k\|^2}{r_k^2}\Big)^{1/2}
      =\Big(1-\frac{T_k}{L(\pvmu_k)}\Big)^{1/2}\quad(r_k>0),\qquad \theta_k=0\quad(r_k=0);
    \end{equation}
    the two forms agree because $T_k/L(\pvmu_k)=\frac{\beta_k}{2}\|\pvDelta_k\|^2\cdot2\sigma_y^2/r_k^2$.

    \begin{lemma}[Verified residual contraction]\label{lem:residual}
    Without any assumption, $r(\pvz^+_k)^2\le r_k^2-\beta_k\sigma_y^2\|\pvDelta_k\|^2$. Hence $\theta_k\in[0,1]$ is well
    defined and $r(\pvz^+_k)\le\theta_kr_k\le r_k$.
    \end{lemma}

    \begin{proof}
    \emph{Step 1.} By \eqref{eq:th_cert}, $\Phi_k(\pvz^+_k)\le L(\pvmu_k)$, i.e.\
    $L(\pvz^+_k)+\frac{\beta_k}2\|\pvDelta_k\|^2\le L(\pvmu_k)$.
    \emph{Step 2.} Multiply by $2\sigma_y^2>0$ and use $2\sigma_y^2L(\pvz)=r(\pvz)^2$:
    $r(\pvz^+_k)^2+\beta_k\sigma_y^2\|\pvDelta_k\|^2\le r_k^2$, which is the first claim.
    \emph{Step 3 ($r_k>0$).} Divide by $r_k^2$: $0\le r(\pvz^+_k)^2/r_k^2\le1-\beta_k\sigma_y^2\|\pvDelta_k\|^2/r_k^2\le1$. So the radicand in
    \eqref{eq:th_theta} lies in $[0,1]$, $\theta_k\in[0,1]$ is well defined, $\theta_k^2\ge r(\pvz^+_k)^2/r_k^2$, and, taking square roots of
    nonnegative numbers, $r(\pvz^+_k)\le\theta_kr_k\le r_k$.
    \emph{Step 4 ($r_k=0$).} Step 2 gives $r(\pvz^+_k)^2+\beta_k\sigma_y^2\|\pvDelta_k\|^2\le0$ with both terms nonnegative, so
    $r(\pvz^+_k)=0$ and $\pvDelta_k=\mathbf 0$; with $\theta_k=0$, $r(\pvz^+_k)=0=\theta_kr_k=r_k$.
    \end{proof}

    \begin{proposition}[Conditional PSNR condition]\label{prop:cond_quality}
    Let Assumptions~\ref{asm:th_noise} and~\ref{asm:th_stability} hold at step $k$, and $s_k=\theta_kr_k$.
    \begin{enumerate}
    \item[(a)] $\|\tilde\pvx_k-\pvx^\star\|\le(s_k+\delta)/m_k$; so, for $d_x$ image entries and peak-to-peak range $R_{\rm pp}$,
      $\mathrm{PSNR}(\tilde\pvx_k,\pvx^\star)\ge10\log_{10}\big(d_xR_{\rm pp}^2m_k^2/(s_k+\delta)^2\big)$.
    \item[(b)] $\|\pvx_k-\pvx^\star\|\ge(r_k-\delta)_+/M_k$; so if $s_k+\delta\le(m_k/M_k)(r_k-\delta)_+$, then
      $\|\tilde\pvx_k-\pvx^\star\|\le\|\pvx_k-\pvx^\star\|$ and the correction does not lower PSNR.
    \end{enumerate}
    \end{proposition}

    \begin{proof}
    \emph{(a).} By the lower inequality of Assumption~\ref{asm:th_stability}, the triangle inequality, $A(\tilde\pvx_k)=A(G(\pvz^+_k))=F(\pvz^+_k)$,
    Lemma~\ref{lem:residual} and Assumption~\ref{asm:th_noise},
    \[
      m_k\|\tilde\pvx_k-\pvx^\star\|\le\|A(\tilde\pvx_k)-A(\pvx^\star)\|\le\|F(\pvz^+_k)-\pvy\|+\|\pvy-A(\pvx^\star)\|
      =r(\pvz^+_k)+\|\pvn\|\le s_k+\delta .
    \]
    Divide by $m_k>0$. With $\mathrm{MSE}=\|\pvx-\pvx^\star\|^2/d_x$, $\mathrm{PSNR}(\pvx,\pvx^\star)=10\log_{10}(d_xR_{\rm pp}^2/\|\pvx-\pvx^\star\|^2)$,
    which decreases in $\|\pvx-\pvx^\star\|$; inserting the bound gives the PSNR bound.
    \emph{(b).} By the triangle inequality, the upper inequality of Assumption~\ref{asm:th_stability} and Assumption~\ref{asm:th_noise},
    $r_k=\|A(\pvx_k)-\pvy\|\le\|A(\pvx_k)-A(\pvx^\star)\|+\|A(\pvx^\star)-\pvy\|\le M_k\|\pvx_k-\pvx^\star\|+\delta$, so
    $\|\pvx_k-\pvx^\star\|\ge(r_k-\delta)_+/M_k$. If $s_k+\delta\le(m_k/M_k)(r_k-\delta)_+$, then by (a)
    $\|\tilde\pvx_k-\pvx^\star\|\le(s_k+\delta)/m_k\le(r_k-\delta)_+/M_k\le\|\pvx_k-\pvx^\star\|$, and PSNR is decreasing in the error norm.
    \end{proof}

    \begin{theorem}[Conditional trajectory error recursion]\label{thm:cond_recursion}
    Let Assumption~\ref{asm:th_noise} hold and, for every $k<T$ with $T\le K-1$, Assumptions~\ref{asm:th_stability}
    and~\ref{asm:th_transition}. Then for $k<T$
    \begin{equation}\label{eq:th_onestep}
      \|\tilde\pvx_k-\pvx^\star\|\le\frac{\theta_kM_k}{m_k}\,e_k+\frac{(1+\theta_k)\delta}{m_k},
      \qquad
      e_{k+1}\le\omega_ke_k+\nu_k,
    \end{equation}
    with $\omega_k=c_{k+1}\theta_kM_k/m_k$ and $\nu_k=c_{k+1}(1+\theta_k)\delta/m_k+b_{k+1}$, and, with empty products equal
    to one,
    \begin{equation}\label{eq:th_unroll}
      e_T\le\Big(\prod_{k<T}\omega_k\Big)e_0+\sum_{j<T}\Big(\prod_{r=j+1}^{T-1}\omega_r\Big)\nu_j .
    \end{equation}
    \end{theorem}

    \begin{proof}
    \emph{Step 1.} The display in the proof of Proposition~\ref{prop:cond_quality}(a) gives $\|\tilde\pvx_k-\pvx^\star\|\le(\theta_kr_k+\delta)/m_k$.
    \emph{Step 2.} The proof of Proposition~\ref{prop:cond_quality}(b) gives $r_k\le M_ke_k+\delta$.
    \emph{Step 3.} Since $\theta_k\ge0$, Step 2 gives $\theta_kr_k\le\theta_kM_ke_k+\theta_k\delta$; inserting this in Step 1,
    $\|\tilde\pvx_k-\pvx^\star\|\le(\theta_kM_ke_k+\theta_k\delta+\delta)/m_k$, the first inequality of \eqref{eq:th_onestep}.
    \emph{Step 4.} By Assumption~\ref{asm:th_transition}, $c_{k+1}\ge0$ and Step 3,
    $e_{k+1}\le c_{k+1}\|\tilde\pvx_k-\pvx^\star\|+b_{k+1}\le c_{k+1}\frac{\theta_kM_k}{m_k}e_k+c_{k+1}\frac{(1+\theta_k)\delta}{m_k}+b_{k+1}=\omega_ke_k+\nu_k$.
    \emph{Step 5 (unrolling).} Induction on $T$. For $T=0$, \eqref{eq:th_unroll} reads $e_0\le e_0$. If it holds for $T<K-1$ and the
    hypotheses hold for $k\le T$, then by Step 4 and $\omega_T\ge0$,
    \[
      \begin{aligned}
      e_{T+1}\le\omega_Te_T+\nu_T&\le\omega_T\Big(\prod_{k<T}\omega_k\Big)e_0+\omega_T\sum_{j<T}\Big(\prod_{r=j+1}^{T-1}\omega_r\Big)\nu_j+\nu_T\\
      &=\Big(\prod_{k<T+1}\omega_k\Big)e_0+\sum_{j<T+1}\Big(\prod_{r=j+1}^{T}\omega_r\Big)\nu_j,
      \end{aligned}
    \]
    the term $j=T$ having the empty product $1$.
    \end{proof}

    \begin{proposition}[Finite-horizon perturbation propagation]\label{prop:cond_coupling}
    Under Assumption~\ref{asm:th_lipschitz} for all $k<T$, with $T\le K$,
    \[
      \|\pvu_T-\pvv_T\|\le\sum_{k<T}\Big(\prod_{r=k+1}^{T-1}\Lambda_r\Big)\|\pvDelta_k\|
      \le\sum_{k<T}\Big(\prod_{r=k+1}^{T-1}\Lambda_r\Big)\,\sigma_k\sqrt{2\,\Delta L_k},
    \]
    where $\Delta L_k=L(\pvmu_k)-L(\pvz^+_k)\ge0$ vanishes at the steps whose commit is the anchor (with $L(\pvmu_k)<\infty$).
    \end{proposition}

    \begin{proof}
    \emph{Step 1.} By \eqref{eq:th_transition} and $\pvmu_k=D_{\sigma_k}(\pvu_k)$,
    $\pvu_{k+1}=\pvz^+_k+\sigma_{k+1}\pveps_k=D_{\sigma_k}(\pvu_k)+\pvDelta_k+\sigma_{k+1}\pveps_k$ (for $k=K-1$ the noise term is absent).
    \emph{Step 2.} The uncorrected chain uses the same $\pveps_k$, so subtracting $\pvv_{k+1}=D_{\sigma_k}(\pvv_k)+\sigma_{k+1}\pveps_k$ gives
    $\pvu_{k+1}-\pvv_{k+1}=D_{\sigma_k}(\pvu_k)-D_{\sigma_k}(\pvv_k)+\pvDelta_k$.
    \emph{Step 3.} By the triangle inequality and Assumption~\ref{asm:th_lipschitz}, $d_k:=\|\pvu_k-\pvv_k\|$ satisfies
    $d_{k+1}\le\Lambda_kd_k+\|\pvDelta_k\|$, with $d_0=0$.
    \emph{Step 4.} The induction of Theorem~\ref{thm:cond_recursion}, Step 5, with $\omega_k=\Lambda_k\ge0$, $\nu_k=\|\pvDelta_k\|$ and
    $e_0=d_0=0$, gives the first inequality.
    \emph{Step 5.} By \eqref{eq:th_cert}, $T_k\le\Delta L_k$, i.e.\ $\|\pvDelta_k\|\le\sigma_k\sqrt{2\Delta L_k}$ since $\beta_k=\sigma_k^{-2}$; multiplying by
    the nonnegative products and summing gives the second. $\Delta L_k\ge0$ by \eqref{eq:th_cert}, and $\Delta L_k=0$ at the steps whose
    commit is the anchor.
    \end{proof}

    Lemma~\ref{lem:traj_kl} compares laws and needs no assumption; Proposition~\ref{prop:cond_coupling} is pathwise but needs
    Assumption~\ref{asm:th_lipschitz}. Neither compares the run with the posterior, and none of these results gives global convergence,
    an unconditional PSNR improvement, or calibrated posterior sampling.

    \subsection{The gated clip derivative}
    \label{app:gate}

    Let the measurement operator clip its output elementwise, $A(\pvx)=\operatorname{clip}(B(\pvx),-1,1)$ with $B$ differentiable, so that $F(\pvz)=A(G(\pvz))=\operatorname{clip}(\mathbf v(\pvz),-1,1)$ with $\mathbf v=B\circ G$, and let $\pvy=A(\pvx^\star)+\pvn$ for the image $\pvx^\star$, with $\pvn\sim\pvN(0,\sigma_n^2I)$ and $\sigma_n>0$, as in the benchmark (noise added after the clip). Let $\tau=\sigma_n\sqrt{2\ln N}$, with $N{>1}$
    the number of coordinates of $\pvy$, and write $c_i=\operatorname{clip}(v_i,-1,1)$. In every gradient of $L$ (the steps and the secants of
    Section~\ref{sec:method_proposal_verification}), the \emph{gated derivative} of the clip passes the incoming gradient where $|v_i|\le1$ and, beyond
    the clip, only where the negative derivative of the misfit with respect to $c_i$ points back into range and the misfit exceeds $\tau$: $v_i>1$ and $c_i-y_i>\tau$, or $v_i<-1$ and $y_i-c_i>\tau$;
    elsewhere it is zero, the exact derivative. The test concerns the scalar signal of coordinate $i$ alone: through the Jacobian of $\mathbf v$, the proximal term and the momentum, a step can still move $v_i$ either way. For fixed $\pvy$ define, for each coordinate $i$,
    \[
      \psi_i(v)=
      \begin{cases}
        \dfrac{(1-y_i)^2}{2\sigma_y^2}+\dfrac{(1-y_i)(v-1)}{\sigma_y^2}, & v>1 \text{ and } y_i<1-\tau,\\[6pt]
        \dfrac{(1+y_i)^2}{2\sigma_y^2}-\dfrac{(1+y_i)(v+1)}{\sigma_y^2}, & v<-1 \text{ and } y_i>-1+\tau,\\[6pt]
        \dfrac{(\operatorname{clip}(v,-1,1)-y_i)^2}{2\sigma_y^2}, & \text{otherwise},
      \end{cases}
    \]
    and $\tilde L(\pvz)=\sum_i\psi_i(v_i(\pvz))$.

    \begin{lemma}[The gated derivative is the gradient of a fixed surrogate]\label{lem:gate_surrogate}
     At every $\pvz$ at which the backward rules of the graph of $\mathbf v$ compute its classical Jacobian and with $|v_i(\pvz)|\neq1$ for all $i$, the gated gradient equals $\nabla\tilde L(\pvz)$.
    
    \end{lemma}

    \begin{proof}
    Write $g_i=(c_i-y_i)/\sigma_y^2$, the derivative of the Gaussian misfit with respect to $c_i$.  If
    $|v_i|<1$, the gate passes $g_i=(v_i-y_i)/\sigma_y^2=\psi_i'(v_i)$. If $v_i>1$, then $c_i=1$ and the gate condition ``$v_i>1$ and $c_i-y_i>\tau$'' reads
    $y_i<1-\tau$, a condition on $y_i$ alone. When it holds the gate passes $g_i=(1-y_i)/\sigma_y^2$, the slope of the first branch; otherwise it passes
    $0$, the slope of the constant third branch. The case $v_i<-1$ is symmetric. The chain rule through $\mathbf v(\pvz)$, which the backward pass computes at such $\pvz$, gives the claim.

    \end{proof}

    \begin{proposition}[Releases from the true side of the clip]\label{prop:gate_release}
    Call coordinate $i$ truly saturated above if $B_i(\pvx^\star)>1$ and below if $B_i(\pvx^\star)<-1$. For every iterate, whatever its dependence on
    $\pvy$, the gate releases a coordinate truly saturated above from the upper side only if $n_i<-\tau$, and one truly saturated below from the lower
    side only if $n_i>\tau$. Hence the number $R$ of coordinates that can be released from the side on which they truly lie satisfies
    $\mathbb E{}_{\pvn}[R]\le N_{\rm sat}\,Q(\sqrt{2\ln N})\le1/\sqrt{4\pi\ln N}$, where $N_{\rm sat}\le N$ is the number of truly saturated coordinates and $Q$ the
    standard Gaussian tail; the expectation is over the measurement noise $\pvn$.
    \end{proposition}

    \begin{proof}
    A coordinate truly saturated above has $y_i=1+n_i$ (since $A_i(\pvx^\star)=1$). By the proof of Lemma~\ref{lem:gate_surrogate}, whenever the iterate lies above the upper clip ($v_i(\pvz)>1$) it is released on the upper side if and only if
    $y_i<1-\tau$, that is $n_i<-\tau$, an event of the noise alone; the lower side is symmetric. The coordinates that can be released from their true side
    therefore form a subset of $S=\{i\text{ above}: n_i<-\tau\}\cup\{i\text{ below}: n_i>\tau\}$, which does not depend on the iterates, and
    $\mathbb E|S|=N_{\rm sat}\,Q(\tau/\sigma_n)=N_{\rm sat}\,Q(\sqrt{2\ln N})$. With  $\varphi(t)=e^{-t^2/2}/\sqrt{2\pi}$, for \(t>0\) we have \(Q(t)=\int_t^\infty\varphi(s)\,ds\le\int_t^\infty\frac{s}{t}\varphi(s)\,ds=\varphi(t)/t\), because \(s/t\ge1\) on \([t,\infty)\) and \(\int_t^\infty s\varphi(s)\,ds=\varphi(t)\), since \(\varphi'(s)=-s\varphi(s)\); at  $t=\sqrt{2\ln N}$,
    $Q(t)\le1/(N\sqrt{2\pi}\sqrt{2\ln N})$, so $\mathbb E|S|\le N_{\rm sat}/(N\sqrt{4\pi\ln N})\le1/\sqrt{4\pi\ln N}$.
    \end{proof}

    The gate changes only the field the steps follow: $\pvg_L$ is Borel (its mask is a comparison), the verifier and the restart test use the exact
    $\Phi_k$, and every result of Sections~\ref{app:theory_setting}--\ref{app:theory_cond} holds verbatim.

\section{Experimental details}
\label{app:exp_detail}

\subsection{Hyperparameter settings}

Unless otherwise specified, natural-image experiments use batch size \(1\),
\(100\) evaluation images, likelihood scale
\(\sigma_y=0.05\), equal to the standard deviation \(\sigma_n=0.05\) of the additive Gaussian measurement noise (added after the clip for clipping operators), which is also the \(\sigma_n\) of the gate threshold. Pixel-space
experiments use an EDM~\citep{karras2022elucidating} noise schedule with
polynomial-7 discretization, \(\sigma_{\max}=80\), \(\sigma_{\min}=0.02\) (the end point of the interpolation, which the \(K\) positive levels do not reach; Appendix~\ref{app:theory_setting}), and \(\sigma_{\mathrm{floor}}=10^{-3}\). Latent-space experiments use noise-level endpoints
\(\sigma_{\max}=10\) and \(\sigma_{\min}=0.02\). The task-specific values of
\(K\) and \(J\) are listed
in Table~\ref{tab:task_specific_hparams}; the step size and momentum are computed at every step from the measured curvature (Section~\ref{sec:method_proposal_verification}). Correction is active at every
reverse step.  We use the
Gaussian likelihood for all tasks.

For MRI reconstruction, we follow the accelerated multi-coil knee MRI protocol of InverseBench~\cite{zheng2025inversebench}, using its pretrained diffusion checkpoint and 96 test samples. For both $\times 4$ and $\times 8$ acceleration factors, we use $K=105$, $J=3$, and $\sigma_y=0.02$. We use the same EDM polynomial-7 noise schedule with $\sigma_{\max}=80$, $\sigma_{\min}=0.02$, and $\sigma_{\mathrm{floor}}=10^{-3}$.

\begin{table}[t]
\centering
\small
\caption{
\textbf{Task-specific settings for pixel-space and latent-space \textsc{PIVOT}.}
Here, \(K\) denotes the number of reverse diffusion steps and \(J\) denotes the
number of gradient evaluations per step (\(J-1\) heavy-ball steps); each value is shared by FFHQ and ImageNet.}
\label{tab:task_specific_hparams}
\begin{tabular}{lrr}
\toprule
Task & $K$ & $J$ \\
\midrule
\multicolumn{3}{l}{\textbf{Pixel-space}} \\
Super resolution $4{\times}$ & 40 & 15 \\
Inpaint (box) & 90 & 15 \\
Inpaint (random) & 100 & 15 \\
Gaussian deblurring & 20 & 140 \\
Motion deblurring & 40 & 20 \\
Phase retrieval & 180 & 70 \\
Nonlinear deblurring & 120 & 120 \\
High dynamic range & 40 & 20 \\
\midrule
\multicolumn{3}{l}{\textbf{Latent-space}} \\
Super resolution $4{\times}$ & 50 & 24 \\
Inpaint (box) & 100 & 24 \\
Inpaint (random) & 50 & 20 \\
Gaussian deblurring & 50 & 30 \\
Motion deblurring & 50 & 30 \\
Phase retrieval & 70 & 35 \\
Nonlinear deblurring & 70 & 80 \\
High dynamic range & 20 & 100 \\
\bottomrule
\end{tabular}
\end{table}

\subsection{Baseline Details}
\label{appendix:baselines}

For DAPS\footnote{\tiny{\url{https://github.com/zhangbingliang2019/DAPS}}}~\cite{zhang2025improving}, DMPlug\footnote{\tiny{\url{https://github.com/sun-umn/DMPlug}}}~\cite{wang2024dmplug},
SITCOM\footnote{\tiny{\url{https://github.com/sjames40/SITCOM}}}~\cite{alkhouri2025sitcom}, ReSample\footnote{\tiny{\url{https://github.com/soominkwon/resample}}}~\cite{song2024solving},
FAST-DIPS\footnote{\tiny{\url{https://github.com/ququlza/FAST-DIPS}}}~\cite{kim2026fastdips}, DCDP\footnote{\tiny{\url{https://github.com/morefre/decoupled-data-consistency-with-diffusion-purification-for-image-restoration}}}~\cite{li2024decoupled}, and
PSLD\footnote{\tiny{\url{https://github.com/LituRout/PSLD}}}~\cite{rout2023solving}, we used the official implementations.
Unless explicitly stated otherwise, each baseline was run with the default
configuration described in the corresponding paper and released codebase.

\paragraph{DCDP~\cite{li2024decoupled}.}
We used the official DCDP implementation. For noisy linear tasks, we used the settings in
Table~\ref{tab:dcdp_hparams}. For nonlinear tasks, we used the default
configuration.
\begin{table}[!htp]
\centering
\caption{\textbf{DCDP hyperparameters for noisy linear tasks.}}
\label{tab:dcdp_hparams}
\setlength{\tabcolsep}{5pt}
\begin{tabular}{lccccc}
\toprule
\textbf{Param (DCDP)} & \textbf{SR} & \textbf{BIP} & \textbf{RIP} & \textbf{GB} & \textbf{MB} \\
\midrule
$t_k$    & 100      & 100      & 100      & 100      & 100 \\
$\alpha$ & $10^{2}$ & $10^{3}$ & $10^{3}$ & $10^{5}$ & $10^{5}$ \\
\bottomrule
\end{tabular}
\end{table}

\paragraph{PSLD~\cite{rout2023solving}.}
We used the official PSLD implementation. The task-specific hyperparameters are
summarized in Table~\ref{tab:psld_hparams}.
\begin{table}[!htp]
\centering
\caption{\textbf{PSLD hyperparameters used in our experiments.}}
\label{tab:psld_hparams}
\setlength{\tabcolsep}{5pt}
\begin{tabular}{lccccc}
\toprule
\textbf{Param (PSLD)} & \textbf{SR} & \textbf{BIP} & \textbf{RIP} & \textbf{GB} & \textbf{MB} \\
\midrule
$\eta$   & $1.0$ & $0.1$     & $0.1$     & $0.1$     & $0.1$ \\
$\gamma$ & $0.1$ & $10^{-3}$ & $10^{-2}$ & $10^{-3}$ & $10^{-2}$ \\
\bottomrule
\end{tabular}
\end{table}
\section{Additional Experiments}
\label{app:add_exp}

\begin{table*}[t]
\centering
\small
\setlength{\tabcolsep}{4pt}
\renewcommand{\arraystretch}{1.08}
\caption{
\textbf{Component ablation of \textsc{PIVOT}.}
Each variant modifies a single design choice.
Results are averaged over 30 images per task, with phase retrieval
following the best-of-four protocol.
}
\label{tab:ablation}
\begin{tabular}{lcccccc}
\toprule
\multirow{2}{*}{\textbf{Variant}}
& \multicolumn{3}{c}{ImageNet latent $4\times$ SR}
& \multicolumn{3}{c}{FFHQ pixel phase retrieval} \\
\cmidrule(lr){2-4}
\cmidrule(lr){5-7}
& PSNR $\uparrow$ & SSIM $\uparrow$ & LPIPS $\downarrow$
& PSNR $\uparrow$ & SSIM $\uparrow$ & LPIPS $\downarrow$ \\
\midrule
\textbf{Full \textsc{PIVOT}}
& 27.364 & 0.744 & 0.384
& 31.572 & 0.886 & 0.197 \\
\midrule
Gradient proposal
& 26.291 & 0.695 & 0.438
& 27.391 & 0.807 & 0.270 \\
Without curvature decay
& 27.126 & 0.733 & 0.396
& 31.869 & 0.890 & 0.193 \\
Polyak first step
& 26.779 & 0.731 & 0.403
& 30.949 & 0.878 & 0.201 \\
Without final verification
& 27.364 & 0.744 & 0.384
& 31.572 & 0.886 & 0.197 \\
Measured-curvature restart
& 27.226 & 0.744 & 0.383
& 16.228 & 0.486 & 0.641 \\
\bottomrule
\end{tabular}
\end{table*}

\subsection{Ablation details}
\label{app:component_ablation}

Table~\ref{tab:ablation} compares the full method with five variants
on ImageNet latent-space $4\times$ super-resolution and FFHQ
pixel-space phase retrieval. Each variant changes one aspect of
the algorithm while keeping the remaining settings fixed.
We evaluate 30 images per task, using the best-of-four protocol
for phase retrieval.

Replacing the inertial update with a gradient proposal
($\alpha=1/M$, $\gamma=0$ at every inner step) reduces PSNR by
1.07 and 4.18\,dB on super-resolution and phase retrieval,
respectively. Applying the Polyak step size from the first
inner step, instead of initializing with $1/M$, reduces PSNR
by 0.59 and 0.62\,dB. Both variants also worsen SSIM and LPIPS
on both tasks.

Removing curvature decay reduces super-resolution PSNR by
0.24\,dB but improves phase-retrieval PSNR by 0.30\,dB,
with consistent trends in SSIM and LPIPS. Retaining larger
curvature estimates can yield smaller subsequent step sizes,
which may explain the improvement in phase retrieval.
For the restart variant, the working curvature is reset to
twice the maximum of the incoming estimate and the observed
secants, rather than doubling its current value.
This reduces phase-retrieval PSNR from 31.57 to 16.23\,dB,
while producing only small, mixed changes in super-resolution.
The reset may discard curvature increases accumulated through
repeated restarts, weakening their stabilizing effect.

Removing final verification commits the final proposal without
comparison against the denoiser anchor. No final proposal was
rejected in the full-method trajectories on either task,
resulting in identical reconstruction metrics.

\begin{figure}[t]
    \centering
    \includegraphics[width=\linewidth]{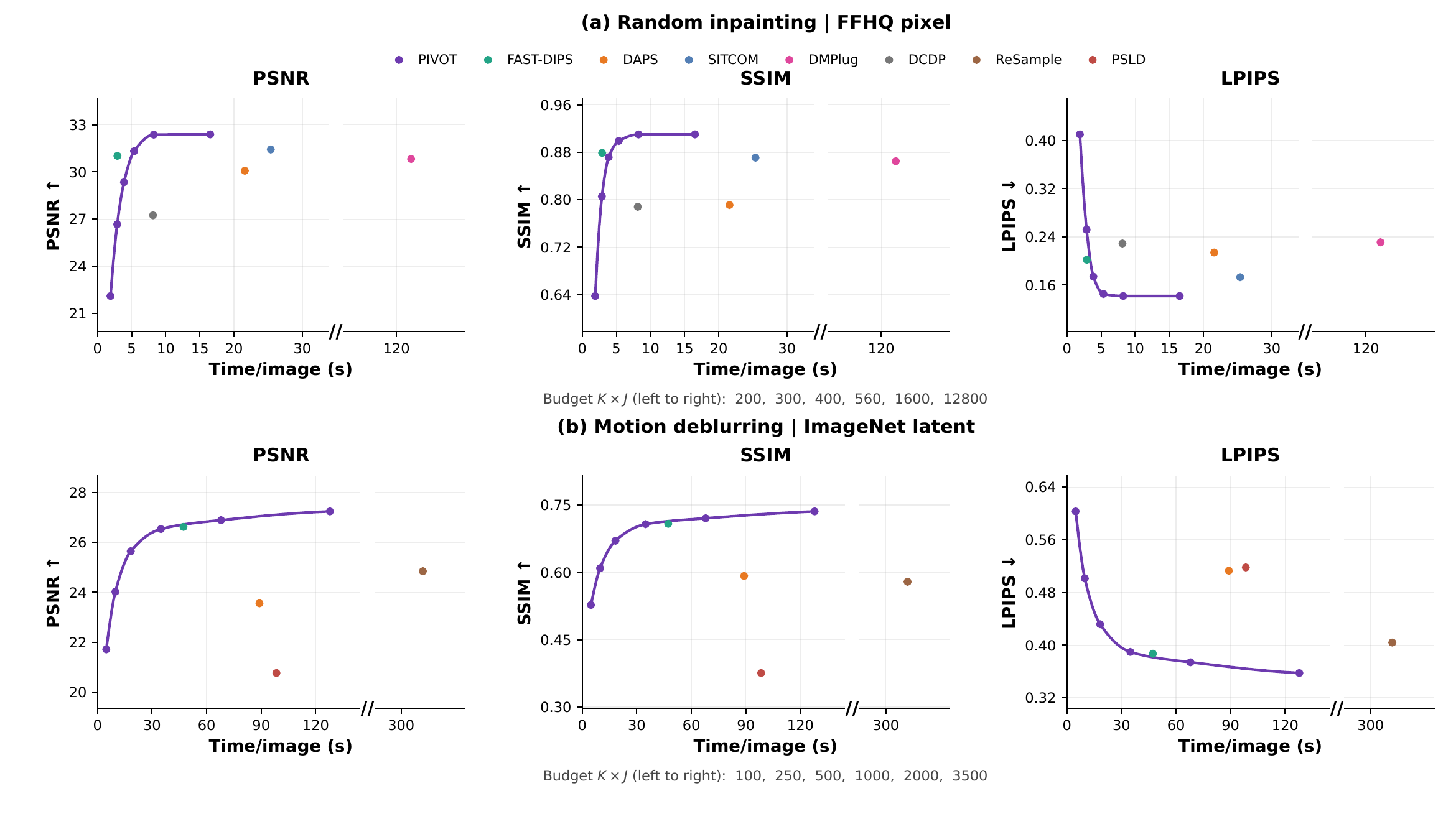}
    \caption{
    \textbf{Speed--quality trade-off.}
    Quality metrics versus runtime as the \textsc{PIVOT} budget
    $K{\times}J$ is varied. Each point represents an average over
    100 samples. Baselines are shown as fixed operating points.
    }
    \label{fig:speed_quality_tradeoff}
\end{figure}

\begin{table}[t]
\centering
\small
\setlength{\tabcolsep}{8pt}
\renewcommand{\arraystretch}{1.08}
\caption{
\textbf{\textsc{PIVOT} budgets used in the speed--quality study.}
}
\label{tab:speed_quality_budgets}
\begin{tabular}{lccc}
\toprule
Task & $K$ & $J$ & $K{\times}J$ \\
\midrule
\multirow{6}{*}{\shortstack[l]{Random inpainting\\(FFHQ pixel)}}
& 20 & 10  & 200   \\
& 30 & 10  & 300   \\
& 40 & 10  & 400   \\
& 56 & 10  & 560   \\
& 80 & 20  & 1600  \\
& 80 & 160 & 12800 \\
\midrule
\multirow{6}{*}{\shortstack[l]{Motion deblurring\\(ImageNet latent)}}
& 50  & 2  & 100  \\
& 50  & 5  & 250  \\
& 50  & 10 & 500  \\
& 50  & 20 & 1000 \\
& 50  & 40 & 2000 \\
& 100 & 35 & 3500 \\
\bottomrule
\end{tabular}
\end{table}

\subsection{\texorpdfstring{Speed--quality trade-off}{Speed--quality trade-off}}
To examine the practical speed--quality behavior of \textsc{\textsc{PIVOT}}, we vary the nominal
computation budget $K{\times}J$, where $K$ is the number of reverse diffusion
steps and $J$ is the number of data-gradient evaluations per step ($J-1$ proposal updates), while keeping all
other parameters fixed within each task. Table~\ref{tab:speed_quality_budgets}
lists the budgets used for FFHQ pixel-space random inpainting and ImageNet
latent-space motion deblurring, and
Figure~\ref{fig:speed_quality_tradeoff} shows the corresponding quality--runtime
curves. On both tasks, increasing the budget improves PSNR, SSIM, and LPIPS,
but the improvements become more gradual as the budget grows, indicating diminishing
returns from additional reverse or proposal steps. These results show that
\textsc{\textsc{PIVOT}} provides a controllable quality--runtime trade--off: moderate budgets
already recover most of the quality achieved in the sweep, while larger budgets can be used
when additional reconstruction accuracy is preferred over runtime. Compared with
representative training-free diffusion solvers at fixed operating points, the resulting
curves give a direct view of the operating regimes in which \textsc{\textsc{PIVOT}}
is Pareto-competitive rather than relying on a single fixed configuration.

\subsection{GPU memory usage}
\label{app:gpu_memory}

We measured peak allocated GPU memory during sampling for each sample
and averaged the results over 20 FFHQ samples with batch size 1.
Table~\ref{tab:peak_memory} reports the results for pixel-space
super resolution and nonlinear deblurring.
\textsc{PIVOT}, FAST-DIPS, DAPS, and DCDP show relatively small differences
in memory usage on both tasks, whereas SITCOM and DMPlug require
substantially more memory.

The larger memory footprints of SITCOM and DMPlug are primarily
associated with backpropagation through the denoiser.
SITCOM retains intermediate denoiser activations for gradient
computation, while DMPlug backpropagates through three connected
DDIM steps, retaining activations across multiple steps. In contrast, \textsc{PIVOT} performs measurement-based corrections without backpropagating through the denoiser. The same measurement gradients
are also used to estimate local curvature through gradient secants,
from which the proposal step size and momentum are determined.
This requires only scalar curvature estimates rather than Hessian
matrices or second-order differentiation, enabling curvature-adaptive
updates while maintaining memory usage comparable to other
low-memory diffusion inverse solvers.

\begin{table}[t]
\centering
\small
\caption{Per-sample peak allocated GPU memory (GiB) during pixel-space
sampling for super resolution $4\times$ and nonlinear
deblurring, averaged over 20 FFHQ samples. Results were obtained on a single RTX 6000 Ada GPU with a batch size of 1.}
\label{tab:peak_memory}
\begin{tabular}{lcc}
\toprule
Method & Super resolution $4\times$ & Nonlinear deblurring \\
\midrule
\textsc{PIVOT}     & \textbf{0.613} & \textbf{1.045} \\
FAST-DIPS & 0.614 & 1.050 \\
DAPS      & 0.626 & 1.063 \\
SITCOM    & 2.036 & 2.593 \\
DMPlug    & 5.236 & 5.793 \\
DCDP      & 0.723 & 1.155 \\
\bottomrule
\end{tabular}
\end{table}

\section{Limitations and future work}
\label{app:limit_future}

\textsc{PIVOT} is designed as a local objective-verification mechanism rather
than a globally convergent optimizer or an exact posterior sampler. Its
nonincrease property applies only to the frozen clean-space objective defined at
the current reverse step. Since the denoiser anchor, noise level, reverse state,
and corresponding frozen objective change across reverse steps, the analysis
does not imply monotonic improvement of the full sampling trajectory, calibrated
posterior samples, or globally optimal reconstructions.  The conditional results (Appendix~\ref{app:theory_cond}) rest on hypotheses that we do not measure.

As a training-free inverse solver, \textsc{PIVOT} relies on the coverage of the
pretrained diffusion prior rather than task-specific retraining. This makes the
method broadly applicable, but reconstruction quality may degrade when the
target domain, noise statistics, or forward model differ substantially from those
represented by the pretrained prior and benchmark protocol.

The method assumes that the composite measurement map \(F=A\circ G\) is
differentiable wherever proposal gradients are used, except that a clip at the output of \(A\) receives the gated derivative of Section~\ref{sec:method_objective_smoothing}. This covers many imaging
forward operators and latent-diffusion decoders, but nondifferentiable
measurement pipelines require differentiable approximations or alternative
gradient estimators. In latent diffusion, reconstruction quality may also depend
on decoder geometry, since measurement gradients are
applied in latent space and propagated through the decoder.

Computational cost is controlled by the number of reverse steps and the number of gradient evaluations per step \(J\). Verification itself uses only one gradient-free objective
evaluation per step; each gradient evaluation is a forward and backward pass through \(F\), which is expensive when \(F\) includes a decoder, a multi-coil MRI operator, or a nonlinear
measurement model.

Empirically, \textsc{PIVOT} is not uniformly best across all tasks and
metrics. Although it achieves strong reconstruction quality in many
settings, competing methods outperform it on certain tasks in terms
of PSNR, SSIM, LPIPS, or runtime. These results indicate that its
advantages vary across tasks and evaluation metrics.

Future work includes broader uncertainty quantification and extensions to
nondifferentiable or partially known forward models. Further evaluation on
higher-resolution, 3D or volumetric, real-noise, and stronger model-mismatch
settings would clarify the transferability of objective-verified local correction
beyond the benchmarks studied here.

\paragraph{Broader impact.}
Improved diffusion-prior reconstruction may benefit scientific and medical
imaging by reducing sampling time or improving reconstruction quality under
limited measurements. However, diffusion priors can also hallucinate plausible
details that are weakly supported by the measurements, especially under severe
undersampling or model mismatch. This is important for face reconstruction and
medical imaging, where reconstructed details may affect privacy, identity, or
clinical interpretation. The method should therefore be used with measurement
residual checks, domain-specific validation, and caution before any high-stakes
deployment.
\newpage
\begin{figure}
\centering
  \makebox[\textwidth][c]{%
    \includegraphics[width=\linewidth]{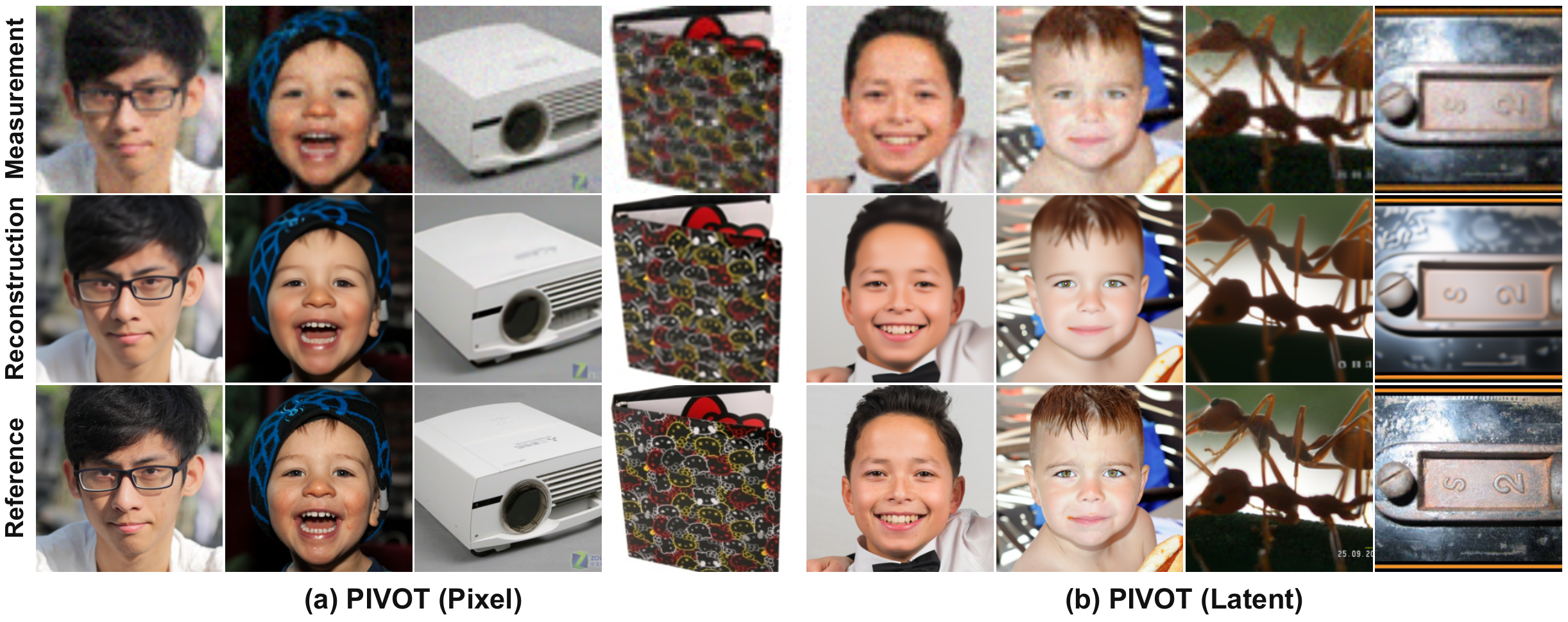}
  }
\caption{Qualitative results for $4\times$ super resolution on FFHQ and ImageNet.}
\label{fig:qual_sr}
\end{figure}

\begin{figure}
\centering
  \makebox[\textwidth][c]{%
    \includegraphics[width=\linewidth]{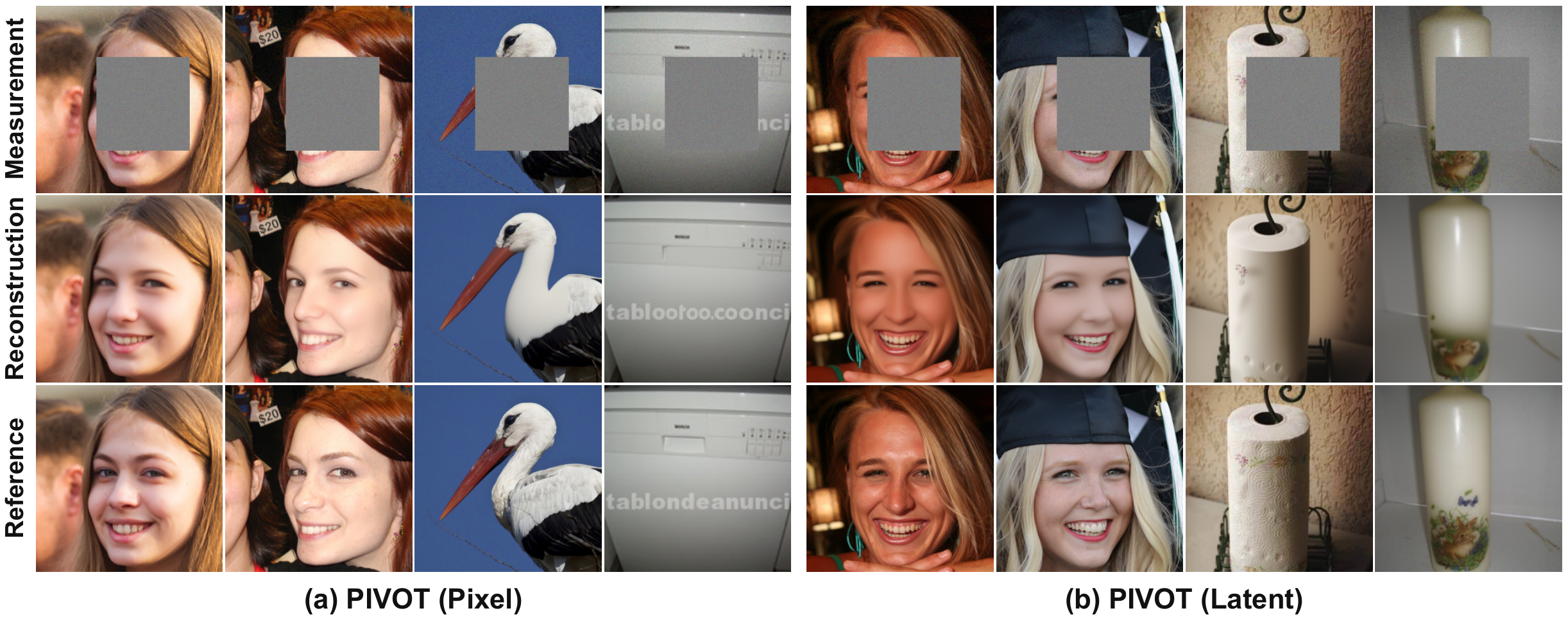}
  }
\caption{Qualitative results for box-mask inpainting on FFHQ and ImageNet.}
\label{fig:qual_box}
\end{figure}

\begin{figure}
\centering
  \makebox[\textwidth][c]{%
    \includegraphics[width=\linewidth]{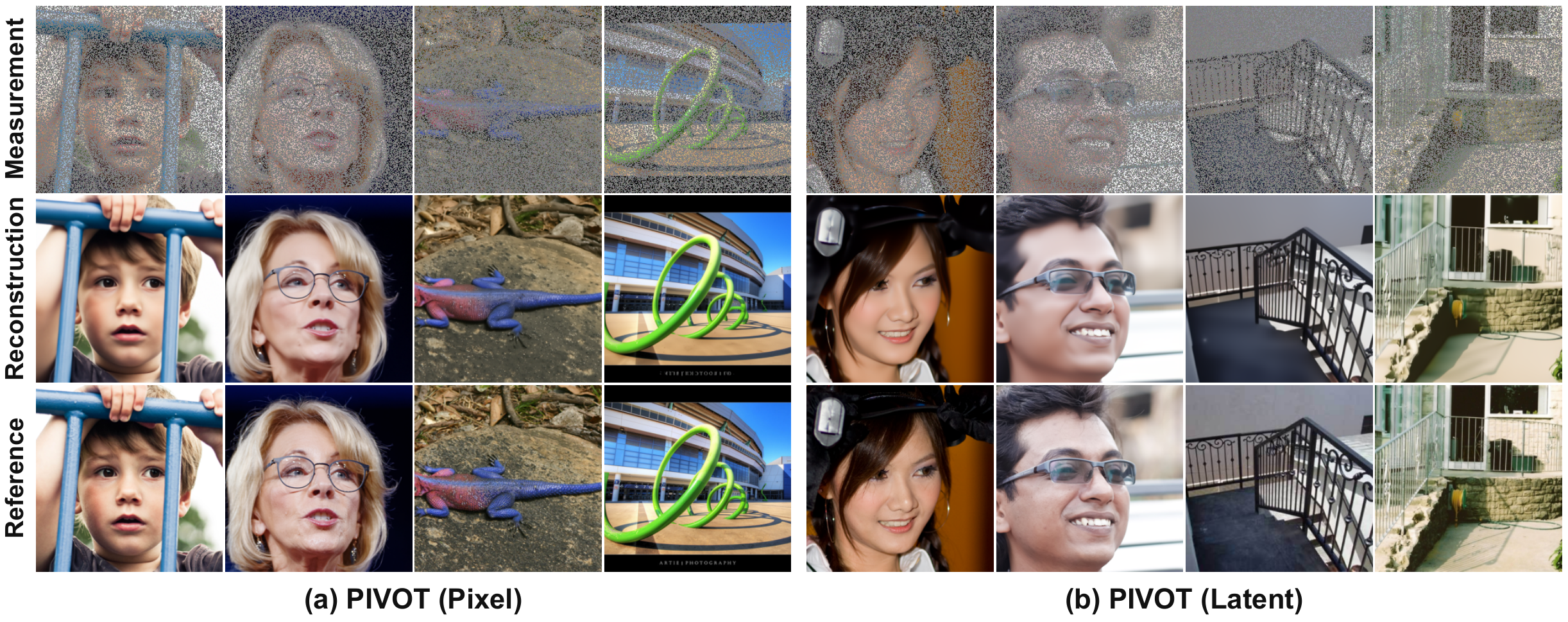}
  }
\caption{Qualitative results for random-mask inpainting on FFHQ and ImageNet.}
\label{fig:qual_random}
\end{figure}

\begin{figure}
\centering
  \makebox[\textwidth][c]{%
    \includegraphics[width=\linewidth]{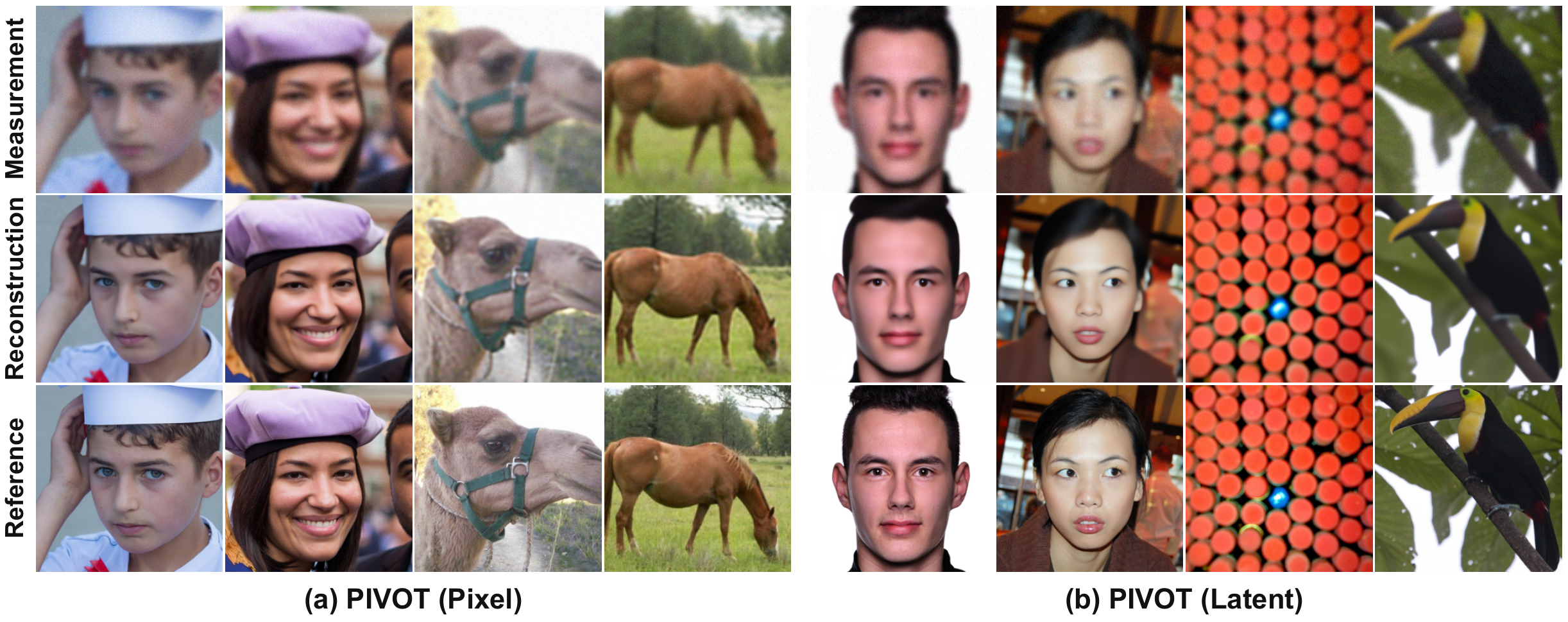}
  }
\caption{Qualitative results for Gaussian deblurring on FFHQ and ImageNet.}
\label{fig:qual_gaussian}
\end{figure}

\begin{figure}
\centering
  \makebox[\textwidth][c]{%
    \includegraphics[width=\linewidth]{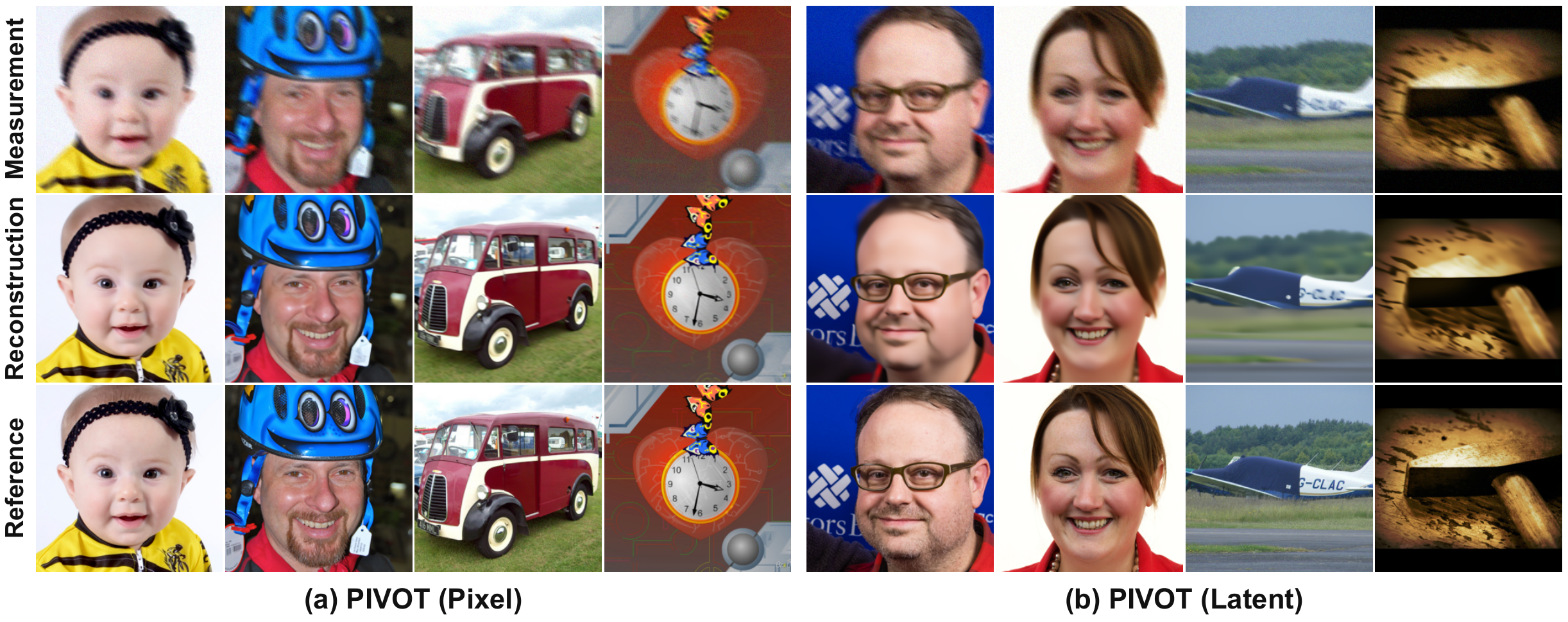}
  }
\caption{Qualitative results for motion deblurring on FFHQ and ImageNet.}
\label{fig:qual_motion}
\end{figure}

\begin{figure}
\centering
  \makebox[\textwidth][c]{%
    \includegraphics[width=\linewidth]{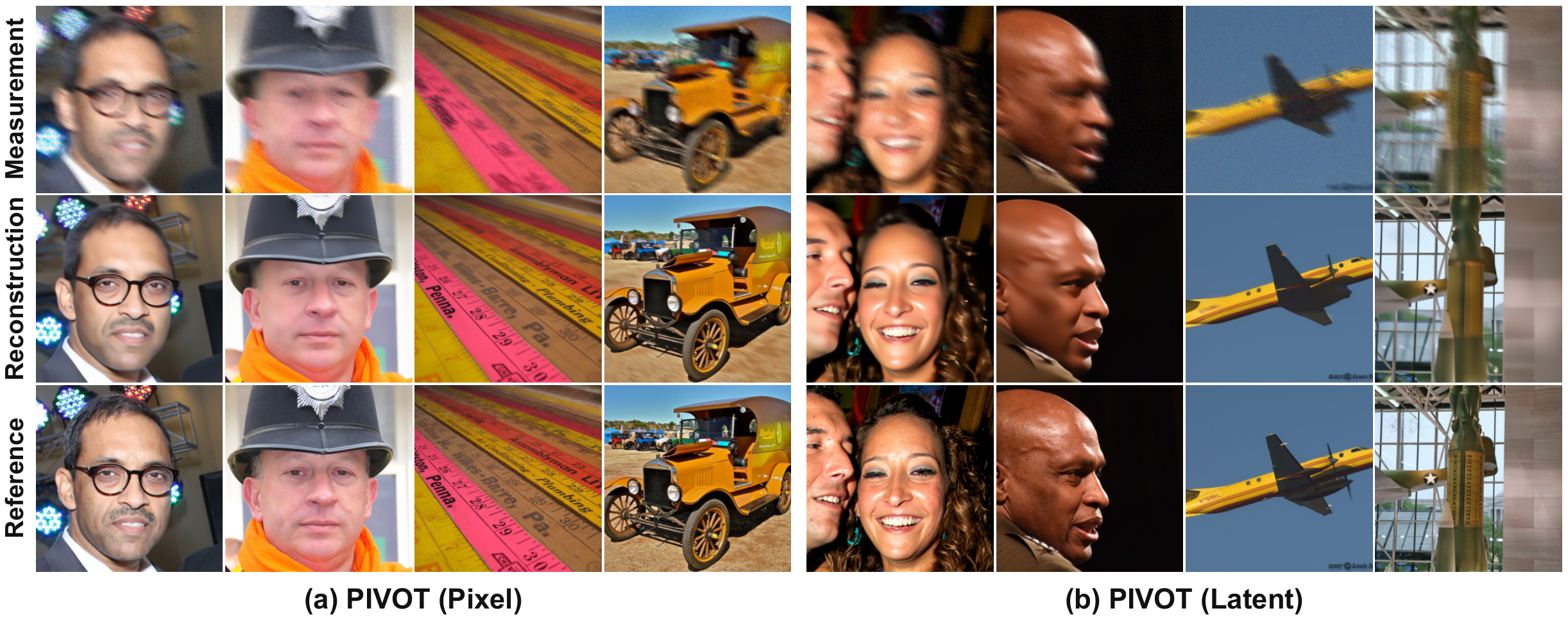}
  }
\caption{Qualitative results for nonlinear deblurring on FFHQ and ImageNet.}
\label{fig:qual_nonlinear}
\end{figure}

\begin{figure}
\centering
  \makebox[\textwidth][c]{%
    \includegraphics[width=\linewidth]{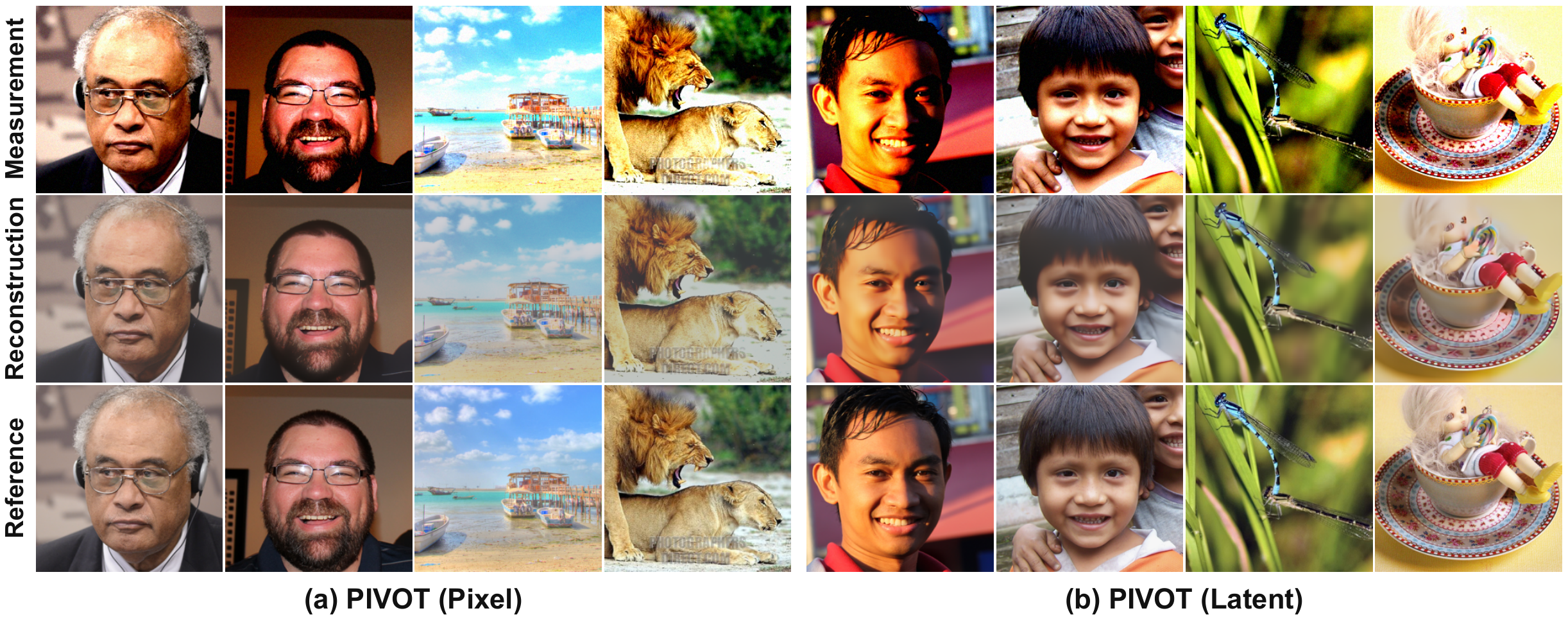}
  }
\caption{Qualitative results for high dynamic range recovery on FFHQ and ImageNet.}
\label{fig:qual_hdr}
\end{figure}

\begin{figure}
\centering
  \makebox[\textwidth][c]{%
    \includegraphics[width=0.8\linewidth]{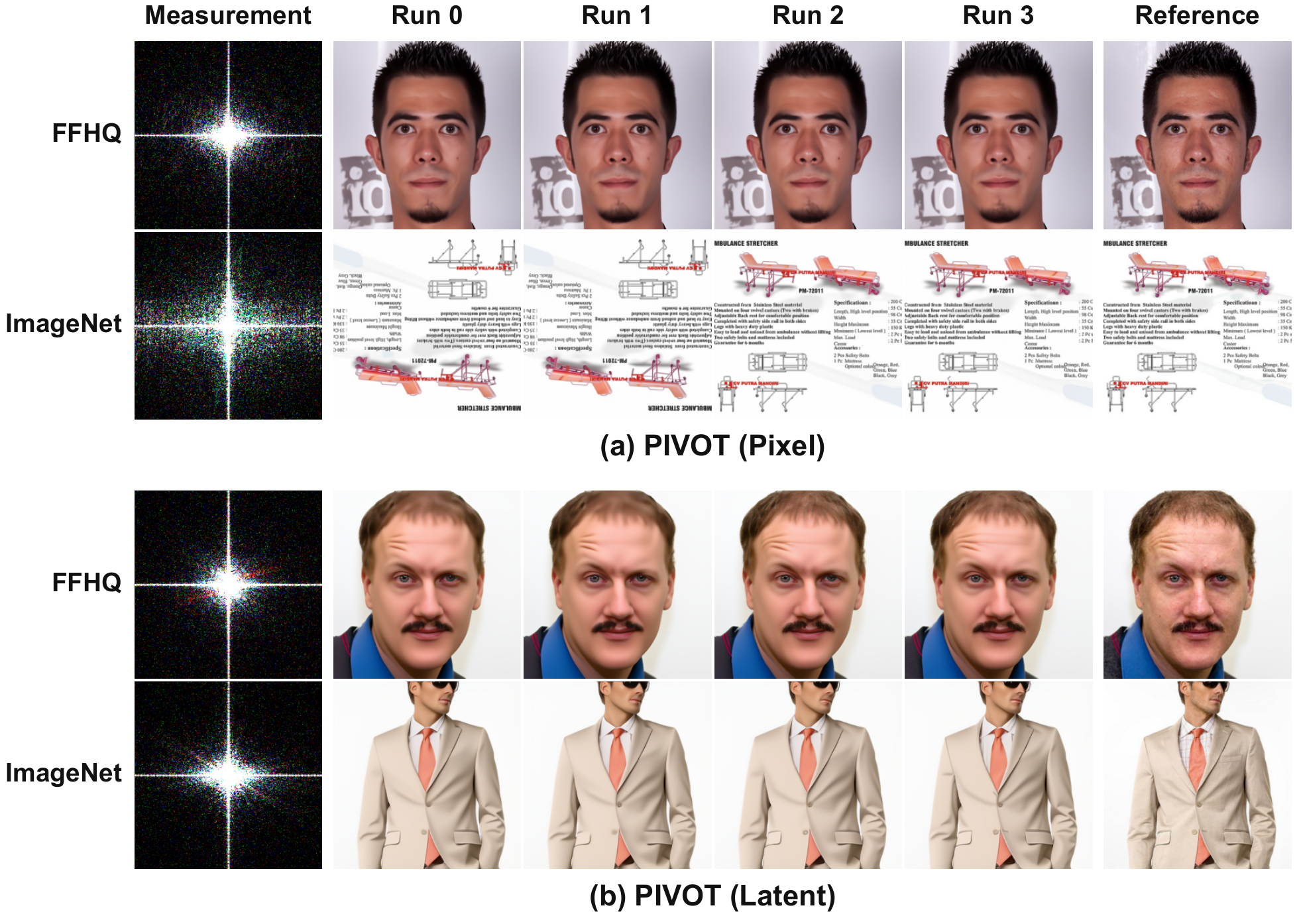}
  }
\caption{Qualitative results for phase retrieval on FFHQ and ImageNet.
For each input, we report reconstructions from four independent trials.}
\label{fig:qual_pr}
\end{figure}

\clearpage

\end{document}